\documentclass{article}

\usepackage{microtype}
\usepackage{graphicx}
\usepackage{subfigure}
\usepackage{afterpage}
\usepackage{adjustbox}
\usepackage{booktabs} 
\usepackage{multicol} 
\usepackage{hyperref}
\usepackage{url}

\usepackage{breakurl}

\usepackage[accepted]{icml2024}
\usepackage{comment}

\usepackage{amsmath}
\usepackage{amssymb}
\usepackage{mathtools}
\usepackage{amsthm}
\usepackage{xcolor}
\usepackage{placeins}
\usepackage{float}
\usepackage[capitalize,noabbrev]{cleveref}

\theoremstyle{plain}

\theoremstyle{definition}

\theoremstyle{remark}

\newcommand{\Tau}{\mathcal{T}}
\usepackage[textsize=tiny]{todonotes}
\usepackage{enumitem}

\icmltitlerunning{A Diffusion Model Framework for Unsupervised Neural Combinatorial Optimization}

\begin{document}

\twocolumn[
\icmltitle{A Diffusion Model Framework for Unsupervised Neural\\ Combinatorial Optimization}



\icmlsetsymbol{equal}{*}

\begin{icmlauthorlist}

\icmlauthor{Sebastian Sanokowski}{JKU,ELLIS}
\icmlauthor{Sepp Hochreiter}{JKU,ELLIS,NXAI}
\icmlauthor{Sebastian Lehner}{JKU,ELLIS}
\end{icmlauthorlist}

\icmlaffiliation{JKU}{Institute for Machine Learning, Johannes Kepler University, Linz, Austria}
\icmlaffiliation{NXAI}{NXAI GmbH}
\icmlaffiliation{ELLIS}{ELLIS Unit Linz}

\icmlcorrespondingauthor{Sebastian Sanokowski}{sebastian.sanokowski@jku.at}

\icmlkeywords{Machine Learning, ICML}

\vskip 0.3in
]



\printAffiliationsAndNotice{}  

\begin{abstract}


Learning to sample from intractable distributions over discrete sets without relying on corresponding training data is a central problem in a wide range of fields, including Combinatorial Optimization. Currently, popular deep learning-based approaches rely primarily on generative models that yield exact sample likelihoods. This work introduces a method that lifts this restriction and opens the possibility to employ highly expressive latent variable models like diffusion models. Our approach is conceptually based on a loss that upper bounds the reverse Kullback-Leibler divergence and evades the requirement of exact sample likelihoods. We experimentally validate our approach in data-free Combinatorial Optimization and demonstrate that our method achieves a new state-of-the-art on a wide range of benchmark problems.\footnote{Code is available at \url{https://github.com/ml-jku/DIffUCO}.}\end{abstract}

\iftrue
\begin{figure*}[ht]

    \centering
    \setlength\tabcolsep{2pt}
  \includegraphics[width=.76\linewidth]{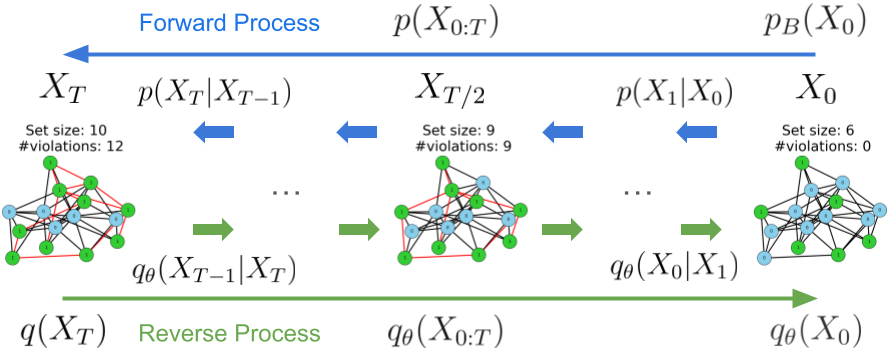}
 
\vspace{-2ex}
\caption{Illustration of DiffUCO's solution generation process on the example of finding the Maximum Independent Set (see Sec.~\ref{sec:benchmarks}) of a graph. Solutions are generated in the reverse diffusion process that is modeled by iteratively sampling the time-conditioned model $q_\theta$.}
\label{fig:illustration}
\end{figure*}
\fi

\section{Introduction}

Sampling from a known but intractable, high-dimensional target distribution like the Boltzmann distribution is of high relevance in many scientific fields like the prediction of molecule configurations \citep{BoltzmannGen}, lattice models in physics \citep{wu_solving_2018} and Monte Carlo integration \citep{NeuralImportanceSampling}. Recently, the works of \citet{hibat-allah_variational_2021} and \citet{VAG-CO} showed that Combinatorial Optimization (CO) can be concisely formulated as such a distribution learning problem where the resulting samples correspond to solutions of a CO problem. In all of these domains, the energy function associated to the distributions of interest is known but obtaining unbiased samples represents a formidable challenge.
Problem areas like CO or lattice models in physics are characterized by discrete target distributions. In these applications, the approximation of the target distribution is predominantly based on products of categorical distributions or autoregressive models.
While product distributions are computationally convenient they lack expressivity due to their inability to represent statistical inter-dependencies. Autoregressive models rely on sequentially generating the components of the samples. For the high-dimensional distributions that are frequently encountered in the aforementioned application areas, this procedure becomes prohibitively expensive.
In addition, there is typically no canonical ordering among the components of samples. Hence, an autoregressive approach appears as an unnatural approach. Intuitively, an issue with the sequential sample generation procedure of autoregressive models is that there is no possibility of correcting sub-optimal decisions once they are made. More formally, it can be shown that autoregressive models are less expressive than energy-based models or latent variable models from a computational complexity point of view \citep{lin2021limitations}.
Approximate likelihood models that utilize latent variables, like Variational Autoencoders (VAE) \citep{VAE} and diffusion models \citep{DicksteinDiff, DenoisingDiffusionModels, SohlDiff}, are trained using data samples and a loss that is based on the Evidence Lower Bound (ELBO). These models have the advantage that they are more expressive and applicable in the discrete setting. However, with these models, it is in general not feasible to evaluate exact sample likelihoods. This problem precludes their application in problems of data-free approximation of target distributions which typically rely on exact sample likelihoods. Recent works that proposed methods to apply diffusion models in data-free approximation of target distributions are limited to the continuous setting \citep{ContDiffModels4, ContDiffModels1, ContDiffModels3, ContDiffModels2}. To the best of our knowledge, the discrete setting remains unexplored.\\
In this work, we propose in Sec.~\ref{sec:novelty} an approach that allows for the application of latent variable models like diffusion models in the problem of data-free approximation of discrete distributions.
We demonstrate our method on paradigmatic problems in this area from the field of CO and obtain state-of-the-art performance over a wide range of benchmarks.
To achieve this our method \emph{Diffusion
for Unsupervised Combinatorial Optimization} (DiffUCO) uses an upper bound on the reverse Kullback-Leibler divergence as a loss function.
We show that the performance of the model consistently improves as we increase the number of diffusion steps used during training. Additionally, we find that during inference the solution quality can be further improved by applying more diffusion steps than during training (see Sec.~\ref{sec:ablations}).
In Sec.~\ref{sec:CE} we propose a significantly more efficient version of a frequently used sampling strategy called \emph{Conditional Expectation}. We show that combined with our diffusion model this method allows for time-efficient generation of high-quality solutions to CO problems. The proposed framework yields a highly efficient and general way of employing latent variable models like diffusion models in the ubiquitous challenge of data-free approximation of discrete distributions.


\section{Problem Description}

\label{sec:descr}
Following \citet{lucas_ising_2014} we will represent CO problems by a corresponding energy function $H: \{ 0,1\}^N \rightarrow \mathbb{R}$ which assigns a scalar value called energy to a given solution $X \in \{ 0,1\}^N$. The dimensionality of $X$ is denoted as $N$ and is called the problem size. The Boltzmann distribution associated to $H$ is defined by:     
\begin{align*}
    p_B(X, \beta) &= \frac{\exp{ (- \beta H(X)) }}{\mathcal{Z}} , \ \ \mathcal{Z} = \sum_X  \exp{ (- \beta H(X)) } ,
\end{align*}
where the parameter $\Tau$ is typically referred to as the temperature and $\beta = 1/\Tau$ as the inverse temperature. In the context of CO the relevant feature of $p_B(X, \beta)$ is that as $\beta \rightarrow \infty$ it places all the probability mass on the solutions with the lowest energy, i.e.~on the \emph{optimal} solutions of $H$. It can be shown that the sample complexity of learning $p_B(X, \beta)$ is in $O(\beta^2)$ \citep{VAG-CO}. Thus, the more probability mass the Boltzmann distribution puts on optimal solutions the harder it becomes to learn.
For large $N$ this distribution is in general intractable, i.e.~sampling from this distribution is difficult for a wide range of frequently encountered energy functions in statistical physics and CO.
Approximate sampling from such an intractable distribution can be realized with Markov Chain Monte Carlo methods but in complex, multi-modal distributions these methods often exhibit insufficient mixing \citep{unbiased1}.
There is a growing interest in training neural networks $q_\theta(X)$ to approximate such a distribution. In these approaches, deep generative models are used to efficiently obtain unbiased samples from Boltzmann distributions (see Sec.~\ref{sec:RelatedWork}). We call the problem of data-free approximation of a target distribution with a deep generative model \emph{Neural Probabilistic Optimization} (NPO).
A common approach to NPO is to minimize the reverse Kullback–Leibler divergence (KL) $D_{KL}(q_\theta(X) \, || \, p_B(X,\beta))$ with respect to the model parameters $\theta$.
Multiplying this objective with $\Tau$ yields an expression that is proportional to the Variational Free Energy $F_\theta(X, T)$, which is given by:
\begin{align}
    L(\theta) &= \mathop{\mathbb{E}}_{X \sim q_{\theta}(X)} [ H(X) + \Tau \log{q_\theta(X)}  ]\\ & = \mathop{\mathbb{E}}_{X \sim q_{\theta}(X)} [ \widehat{F}_\theta(X,\Tau)  ] :=  F_\theta(X,\Tau),
    \label{eq:kl_div}
\end{align}
where $\widehat{F}_\theta(X,\Tau) := H(X) + \Tau \log{q_\theta(X)}$ \citep{wu_solving_2018}.
Minimizing this objective requires exact evaluation of $\log{q_\theta(X)}$ not only to calculate the Variational Free Energy objective but also to backpropagate through the expectation of this objective. This can for example be done with REINFORCE \citep{REINFORCE} gradient estimator:
\vspace{0.0cm}
\begin{align*}
    \nabla_\theta L(\theta)  &= \mathop{\mathbb{E}}_{X \sim q_{\theta}(X)} [ \widehat{F}_\theta(X,\Tau)  \, \nabla_\theta \log{q_\theta(X)}]. 
    \label{eq:back_reinf}
\end{align*}
\vspace{0.0cm}
To minimize this objective it is necessary to choose a generative model for which the sample likelihood $q_{\theta}(X)$ can be efficiently evaluated. Thus, optimizing this objective with a latent variable model like VAEs or diffusion models is not possible since these models do not allow an exact evaluation of $q_{\theta}(X)$.  In Sec.~\ref{sec:joint_diff-obj}, we propose an objective based on an upper bound of the reverse KL divergence that mitigates this issue. 
In principle, also other divergences could be used to approximate a given target distribution. For example, both the reverse and the forward KL divergence are special cases of Renyi divergences \citep{RenyiDivergence} which all require exact evaluation of $\log{q_\theta(X)}$. By introducing an upper bound on the divergence as we do in Sec.~\ref{sec:joint_diff-obj}, the evaluation of $\log{q_\theta(X)}$ is avoided.
Which divergence is the most suitable depends on the application as Renyi divergences are either mass covering or mode seeking \citep{minka2005divergence}.
We will focus on the reverse KL divergence and note that our framework can be applied to any other Renyi divergence.

\subsection{Unsupervised Neural Combinatorial Optimization}
\label{sec:neural_CO}
In Combinatorial Optimization (CO) the task is to find the solution $X \in \{ 0, 1 \}^N$ that minimizes an objective $O: \{ 0, 1 \}^N \rightarrow \mathbb{R}$ under the constraint that $X$ is in the set of feasible solutions $f$.
In \citet{lucas_ising_2014} it is shown that many of these CO problems can be described by identifying the objective $O$ with an energy function $H_Q$ in the form of:
\vspace{0.0cm}
\begin{equation}
H_Q(X) = \sum_{i,j} Q_{ij} X_i X_j ,
\label{eq:CUBO}
\end{equation}
\vspace{0.0cm}
Here, $Q_{ij}$ is defined by the problem at hand, and for each CO problem, it is easy to choose $Q \in \mathbb{R}^{N \times N}$ in a way so that global minima correspond to the optimal solutions of the problem \citep{lucas_ising_2014}. In the context of optimization and quantum computing, this kind of energy function is well known as the  Quadratic Unconstrained Binary Optimization (QUBO) formulation of CO problems \citep{QUBO}.\\
In these terms, the goal of UCO is to train a generative model to sample low energy configurations $X$ when conditioned on a CO problem instance $Q$ which is sampled from a distribution $\omega(Q)$. The goal in UCO is to train a model without the use of any example solutions. The corresponding optimization task is to find parameters $\theta$ that minimize the following loss:
\vspace{0.0cm}
\begin{equation}
 L(\theta) = \mathbb{E}_{Q \sim \omega(Q)} \left [ \mathbb{E}_{X \sim q_\theta(X|Q) } [ H_Q(X) ] \right ].
\label{eq:red_obj}
\end{equation}
\vspace{0.0cm}
The square matrix $Q$ can be regarded as the adjacency matrix of a graph with weighted edges, which is why the dependence on $Q$ in $q_\theta(X|Q)$ is often parameterized by a Graph Neural Network (GNN) \citep{cappart_combinatorial_2022}.
Instead of minimizing Eq.~\ref{eq:CUBO} directly, numerous approaches \citep{hibat-allah_variational_2021, sun_annealed_2022,  sanokowski_one_2022, VAG-CO} instead frame this as an NPO problem by minimizing Eq.~\ref{eq:kl_div}. Here, ideas from Simulated Annealing \citep{sim_ann_1983} are borrowed, i.e.~the objective is first optimized at a high temperature and then the temperature is gradually reduced according to a predefined temperature schedule. In the limit of $\Tau \rightarrow 0$, Eq.~\ref{eq:kl_div} reduces to Eq.~\ref{eq:red_obj} (up to the expectation over $\omega(Q)$) which indicates that in this limit the energy of the sampled solutions is minimized without any additional entropy regularization. Consequently, at the end of this annealing procedure, the average energy of the model's samples can be considered as a measure of the residual bias of the learned approximation of the Boltzmann distribution. \citet{VAG-CO} motivate this technique theoretically in the form of a curriculum learning approach and it has been shown empirically \citep{ hibat-allah_variational_2021, sun_annealed_2022, VAG-CO} that this method leads to models with solutions of higher quality.

\section{Neural Probabilistic Optimization Objective for Approximate Likelihood Models}
\label{sec:novelty}
\label{sec:joint_diff-obj}
Training an approximate likelihood model to learn to generate samples from a target distribution\footnote{In contexts, where it is not relevant we suppress the dependence of the Boltzmann distribution on $\beta$ and denote it as $p_B(X)$.} $p_B(X)$
in the data-free setting is a challenging problem because the evaluation of the sample likelihood within the reverse KL divergence is intractable (see Sec.~\ref{sec:RelatedWork}).
In the following, we will derive an objective that can be used with approximate likelihood models like VAEs or diffusion models.
For that consider the task of approximating a target distribution $p_B(X)$ by using the marginal sample probability of a latent variable model $q_\theta(X) = \int q_\theta(X|Z) \, q(Z) \, dZ$, where $q(Z)$ is an easy to sample prior distribution over latent variables $Z$. Due to the intractability of this marginalization it is not possible to directly minimize $D_{KL}(q_\theta(X) \, || \, p_B(X))$. Instead we use a simple tractable upper bound (see  App.~\ref{app:upper_bound} for a derivation) which is given by: 
\vspace{0.0cm}
\begin{equation}
    D_{KL}(q_\theta(X) \, || \, p_B(X)) \leq D_{KL}(q_\theta(X,Z) \, || \, p(X,Z)),
    \label{eq:joint_var_ub}
\end{equation}
\vspace{0.0cm}
where $q_\theta(X,Z) = q_\theta(X|Z) \, q(Z)$ and $p(X,Z) = p(Z|X) \, p_B(X)$. $p(Z|X)$ can either be learned by parameterizing it with a neural network like it is done in a VAE or it can be simply defined in a way so that $p(Z|X)$ maps samples $X \sim p_B(X)$ onto a pre-defined stationary distribution $q(Z)$ as it is done in diffusion models. 
The advantage of this upper bound is that it is tractable as $q_\theta(X,Z)$ and $p(X,Z)$ can be exactly evaluated and samples $ X, Z \sim q_\theta(X,Z)$ are easy to obtain.
Therefore, as we have shown that the reverse KL of approximate likelihood models $D_{KL}(q_\theta(X) \, || \, p_B(X))$ is bounded from above by the reverse joint KL $D_{KL}(q_\theta(X,Z) \, || \, p(X,Z))$, we can decrease the intractable reverse KL divergence by decreasing the tractable reverse joint-KL divergence instead. According to Gibbs' inequality the KL divergence between two distributions is non-negative and equals zero iff the distributions are identical. Thus, if the right-hand side of the inequality is zero, the inequality becomes an equality. In this case, the target distribution is exactly approximated. Otherwise, the right-hand side of the inequality bounds the approximation error given by the KL divergence on the left-hand side.
In the following we call this objective $D_{KL}(q_\theta (X,Z) \, || \, p(X,Z))$ for NPO the \emph{Joint Variational Upper Bound}.
Equation \ref{eq:joint_var_ub} is a special case of the data processing inequality for Renyi divergences \citep{RenyiDivergence} and can therefore be applied analogously to other Renyi divergences.

\subsection{Training Diffusion Models with the Joint Variational Upper Bound}
\label{sec:joint_diff-obj_loss}
To minimize the right-hand side of Eq.~\ref{eq:joint_var_ub}, we train a diffusion model. As illustrated in Fig.~\ref{fig:illustration} the forward diffusion process transforms the target distribution $p_B(X_0)$ into a \emph{stationary distribution} $q(X_T)$ through iterative sampling of a \emph{noise distribution} $p(X_t|X_{t-1})$ for a total of $T$ iterations. The diffusion model is supposed to model the reverse process, i.e.~to map samples $X_T \sim q(X_T)$ to $X_0 \sim p_B(X_0)$ by iteratively sampling $ q_\theta(X_{t-1} | X_t)$. 
The joint distribution of the reverse process is given by $q_\theta(X_{0:T}) = q(X_T) \prod_{t=1}^{T} q_\theta(X_{t-1}|X_t)$ and we can easily obtain samples $X_{0:T} \sim q_\theta(X_{0:T})$.
The joint distribution of the forward process
$p(X_{0:T}) = p_B(X_0) \prod_{t = 1}^T p(X_t|X_{t-1}) $ is analytically known, but in the data-free setting samples $X_{0:T} \sim p(X_{0:T})$ are not available.
We apply the upper bound in Eq.~\ref{eq:joint_var_ub} by substituting $Z$ with $X_{1:T}$ and get the inequality:
\vspace{0.0cm}
\begin{equation*}
   D_{KL}(q_\theta(X_0) \, || \, p_B(X_0)) \leq D_{KL}(q_\theta (X_{0:T}) \, || \, p(X_{0:T})).
\end{equation*}
\vspace{0.0cm}
We use this Joint Variational Upper Bound as a loss function and minimize it with respect to $\theta$ using RAdam \citep{RADAM}. The resulting expression can be simplified (see App.~\ref{app:deriv_obj}) to:
\vspace{0.0cm}
\begin{equation}
\begin{aligned}
    \Tau \, D_{KL}& \left ( q_\theta(X_{0:T}) \, || \, p(X_{0:T}) \right ) = \\ & - \mathcal{T} \cdot \sum_{t = 1}^{T} \mathbb{E}_{X_{T:t} \sim q_\theta(X_{T:t})} \left [ S(q_\theta(X_{t-1}|X_t)) \right ]  \\ & - \mathcal{T} \cdot  \sum_{t=1}^{T} \mathbb{E}_{X_{T:t-1} \sim q_\theta(X_{T:t-1})} \left [\log p(X_{t}|X_{t-1}) \right ] \\ 
       & +  \mathbb{E}_{X_{T:0} \sim q_\theta(X_{T:0})} \left [ H_Q(X_0) \right ]  + C,\\
\end{aligned}
\label{eq:loss}
\end{equation}
\vspace{0.0cm}
where  $S(q_\theta(X_{t-1}|X_t))$ is the Shannon entropy of $q_\theta(X_{t-1}|X_t)$. The logarithm of the normalization constant $\mathcal{Z}$ of the Boltzmann distribution is absorbed into the constant $C$ as it depends neither on $X$ nor on $\theta$.\\
The first term on the right-hand side of Eq.~\ref{eq:loss} represents an entropy regularization that encourages our model to explore the solution space in the initial phase of training, i.e.~when annealing starts with high values of $\Tau$. The second term represents a coupling between the forward diffusion and the reverse diffusion. Essentially, it ensures that the model yields reverse diffusion paths that are likely under the forward diffusion process. The third term represents the CO objective given in Eq.~\ref{eq:red_obj} and the fourth term is a collection of additive constants that are independent of $\theta$.\\
As it is done in \citet{DenoisingDiffusionModels} we use a time-conditioned diffusion model $q_\theta(X_{t-1}|X_t,t)$, where we provide the model with a one-hot encoding of the current time step $t$. In the following, we omit the conditional dependence on $t$ for notational convenience.

\subsection{Noise Distributions}
\label{sec:noise_distr}
Due to the discrete nature of UCO, noise distributions have to be discrete. In this work, we apply the following two noise distributions.

\textbf{Categorical Noise Distribution.}
In \citet{DiscreteDiffusionModels} the categorical distribution is proposed as a noise distribution in discrete diffusion models.\\
In the binary setting $X \in \{ 0,1\}^N$, this distribution is given by
$ p(X_t | X_{t-1}) =  \prod_{i=1}^N p(X_{t,i} | X_{t-1})$. Here, $X_{t,i}$ is the $i$-th component of $X_t = (X_{t,1}, ..., X_{t,N})$ and $p(X_{t,i} | X_{t-1})$ is a Bernoulli distribution which is given by
\vspace{0.0cm}
\begin{equation*}
p(X_{t,i} | X_{t-1}) = \begin{cases}
(1-\beta_t)^{1- X_{t-1,i}} \cdot \beta_t^{X_{t-1,i}}  \ \text{for $X_{t,i} = 0$}\\
(1-\beta_t)^{X_{t-1,i}} \cdot \beta_t^{1-X_{t-1,i}} \  \text{for $X_{t,i} = 1$}
\end{cases}
\end{equation*}
\vspace{0.0cm}
where $\beta_t = \frac{1}{T-t+2}$. Effectively, this distribution flips the value of $X_{t-1,i}$ with a probability of $\beta_t$, independently of the energy function and the value of $X_{t-1,i}$. At $t=T$ this noise distribution yields the uniform distribution which is taken to be the stationary distribution.

\textbf{Annealed Noise Distribution.}
When the Categorical Noise Distribution is used, the second term on the right-hand side of Eq.~\ref{eq:loss} becomes zero at $\Tau = 0$ and does not contribute to training.
Therefore, we propose another noise distribution such that the resulting logarithm of the noise distribution scales inversely with the temperature. That way $\Tau$ cancels out and this loss term does not vanish anymore at $\Tau = 0$.
We use the target distribution at higher temperatures $1/(\beta \beta_t)$ to construct a noise distribution according to $p(X_t|X_{t-1}) = p_B(X_t, \beta \, \beta_t)$, where $\beta_t \in [0,1[$. At $\beta_{T} = 0$ the Boltzmann distribution is equal to a uniform distribution, which is used as the stationary distribution $q(X_T)$. In the forward diffusion process, i.e.~going from $t=0$ to $t=T$, we sequentially decrease $\beta_t$ with a linear schedule according to $\beta_t = 1 - \frac{t}{T}$.
The resulting Annealed Noise Distribution reads:
\begin{equation*}
    p(X_t| X_{t-1}) \propto \exp{(-\beta \, \beta_t \, X_{t}^T Q X_{t})},
\end{equation*}
where the normalization constant of $p_B(X, \beta \, \beta_t)$ is omitted since it is irrelevant to the optimization of the model for the reason that is given in Sec.~\ref{sec:joint_diff-obj_loss}. In contrast to the Categorical Noise Distribution, this noise distribution depends on the energy function of the considered problem. In the forward direction, it initially favors to enter low-energy states and gradually turns into a uniform random selection of solutions. 
An empirical comparison between the Categorical Noise Distribution and the Annealed Noise Distribution is presented in Sec.~\ref{sec:ablations}.

\section{Joint Variational Diffusion Models in Neural Combinatorial Optimization}
\label{sec:methods}
In the following, we discuss the calculation of the individual terms in Eq.~\ref{eq:loss}.\\
Each diffusion step is represented by a conditional mean-field distribution: $$q_\theta(X_{t-1}|X_t) = \prod_i \widehat{q_\theta}(X_t)_i^{X_{t-1,i}} \, (1-\widehat{q_\theta}(X_t)_i)^{1-X_{t-1,i}}, $$ where $\widehat{q_\theta}(X_t)_i := q_\theta(X_{t-1,i} = 1|X_t)_i$. \\ 
With these product distributions, the expectation of $H_Q(X)$ with respect to $q_\theta(X_{t-1}|X_t)$ can be conveniently calculated in closed form (see App.~\ref{app:exact_expectations}):  
\begin{equation*}
\begin{aligned}
    \mathbb{E}_{X_{t-1} \sim q_\theta(X_{t-1}|X_t)} [H_Q(X_{t-1})]  & = \sum_{i \neq j} Q_{ij} \, \widehat{q_\theta}(X_{t})_i \, \widehat{q_\theta}(X_{t})_j \\ &+ \sum_i Q_{ii} \, \widehat{q_\theta}(X_{t})_i.
\end{aligned}
\end{equation*}
Likewise, we can compute the exact expectation of terms that include the noise distribution and the entropy (see App.~\ref{app:exact_expectations}).
The result for the entropy term reads:
\vspace{0.0cm}
\begin{equation*}
\begin{aligned}
    S(q_\theta(X_{t-1}|X_t)) &= -\sum_{i} \left [ \widehat{q_\theta}(X_t)_i \log{\widehat{q_\theta}(X_t)_i} \right. \\ & \left. -  (1-\widehat{q_\theta}(X_t)_i) \log{(1 - \widehat{q_\theta}(X_t)_i)} \right ].
\end{aligned}
\end{equation*}
\vspace{0.0cm}
The ultimate goal of UCO is to train a model on a dataset of CO problem instances $G_m \sim \omega(G_m)$, which can then be used at test time to generate solutions to i.i.d.~CO problems via a time-efficient inference process. In our approach, this is realized by conditioning a diffusion model on each CO problem instance and by using the empirical mean over $M_\omega$ problem instances. We implement such a conditional generative model via the GNN-based architecture described in App.~\ref{app:architecture}. The corresponding overall loss function becomes:
$$L_\theta = \frac{\Tau}{M_\omega} \sum_{m = 1}^{M_\omega} D_{KL}(q_\theta(X_{0:T}| G_{m}) \, || \, p(X_{0:T}|G_m)),$$
where the energy function is contained in the joint-target distribution $p(X_{0:T}|G_m)$ and depends on the specific CO problem instance at hand\footnote{In the following we suppress the conditional dependencies on $G_m$ to keep the notation simple.}. The expectations with respect to $q_\theta(X_{0:T}| G_{m})$ in the loss above are approximated with a Monte Carlo estimate based on $M_{KL}$ samples drawn from the reverse process. We discuss the computation of the gradients of our loss in App.~\ref{app:gradient} and details on the optimizer for the model parameters $\theta$ in App.~\ref{app:hyperparams}.
In all of our experiments, we use annealing, i.e.~we linearly decrease the temperature $\Tau$ from $\Tau_{\mathrm{start}}$ to zero. We show in Sec.~\ref{sec:ablations} that this indeed improves the solution quality of our model.

\subsection{Conditional Expectation}
\label{sec:CE}
Conditional Expectation (CE) is an iterative strategy to obtain samples from a mean-field distribution that have a better-than-average solution quality \citep{raghavan1988probabilistic,karalias_erdos_2020}. In the following, we consider a mean-field distribution $q_\theta(X) = \prod_i q_\theta(X_i)$ that is defined by the vector of Bernoulli probabilities $v$ with components $v_i = q_\theta(X_i)$. In the first step of CE, we sort the components of $v$ in descending order. Then starting from $i = 0$ the $i$-th component of the resulting vector is set to $0$ to obtain a vector $\omega_0$ and set to $1$ to obtain another vector  $\omega_1$. So in this case $\omega_0 = (0, p_1, …, p_N)$ and $\omega_1 = (1, p_1, …, p_N)$, where $p_i$ is the $i$-th component of $v$ after it is sorted. Then $H(\omega_0)$ and $H(\omega_1)$ are computed and $v$ is updated to the configuration $\omega_j$, where $j= \mathop{\mathrm{argmin}}_{l \in \{ 0, 1\}} H(\omega_l)$ and retains the resulting value of $v_i$ in the following steps. Next $i$ is incremented to $ i + 1$ and the process is repeated until all $v_i$ are set to either $0$ or $1$.

\textbf{Tokenized Conditional Expectation.}
In this work we propose to combine CE with Subgraph Tokenization (ST) \citep{VAG-CO}, where $k \in \mathbb{N}^+$ solution variables are grouped together to form a subgraph token. In the case of a token size $k$ there are $2^k$ possible configurations of a given token. The CE procedure described above is carried out analogously by consecutively selecting token configurations $\omega_j$ with $j=\mathop{\mathrm{argmin}}_{l \in \{1, …, 2^k \} } H(\omega_l)$. Then $i$ is incremented to $i + k$ and the next token gets its value assigned. This process above is repeated until all components of $v$ are set. When $k$ is not exceedingly large the values ${H(\omega_l)}$ can be computed in parallel for all of the $2^k$ configurations. This leads to a computational speed up with respect to CE since the number of iterations is reduced by a factor of $k$.
The corresponding efficiency improvement is demonstrated experimentally in Sec.~\ref{sec:benchmarks} and App.~\ref{app:ST_ablation}.

\subsection{EGN is secretly a one-step Joint Variational Diffusion Model}
\label{sec:egn_sec}
The UCO methods Erdős Goes Neural (EGN, \citet{karalias_erdos_2020}) and its annealed variant EGN-Anneal \citep{sun_annealed_2022} use a mean-field distribution $q_\theta(X)$ which is parameterized by a GNN. They also use Eq. \ref{eq:kl_div} as a loss function, which corresponds to minimizing $D_{KL}(q_\theta(X) \, || \,  p_B(X))$. However, due to their reliance on the deterministic CE sampling procedure, their models would only generate one solution, which is why they use additional random node features $Z \sim q(Z)$ so that more than a single solution can be sampled from the models. Consequently, these methods are actually optimizing $\mathbb{E}_{Z \sim q(Z)} [ D_{KL}(q_\theta(X|Z) \, || \,  p_B(X)) ]$ which is equal to $D_{KL}(q_\theta(X,Z) \, || \,  p_B(X)) - \hat{C}$, where $\hat{C} = \int q(Z) \log{q(Z)} d \, Z$ is constant. It can then be shown that:
\begin{align}
\begin{split}
 D_{KL}(q_\theta(X,Z)  \, || \, & p_B(X)) + C \\
& = D_{KL}(q_\theta(X,Z)  \, || \, p(X,Z)).
\end{split}
\label{eq:loss_comp}
\end{align}
The right-hand side of this equation is equivalent to our loss in Eq.\ref{eq:loss} when only one diffusion step is used, i.e.~$T=1$. Finally, since for diffusion models, $C$ and $\hat{C}$ are independent of model parameters (see App.~\ref{app:EGNisOneStepDiffUCO}) the gradients of the left-hand side and right-hand side of the Eq.~\ref{eq:loss_comp} are identical.
Consequently, it can be concluded that DiffUCO generalizes EGN and EGN-Anneal.

\section{Related Work}
\label{sec:RelatedWork}
\textbf{Variational Autoencoders.}
Variational Autoencoders (VAEs) are latent variable models, where samples $X$ are drawn by first sampling latent variables $Z$ from a prior distribution $q(Z)$ which are then processed by a stochastic decoder $q_\theta(X|Z)$ that is parameterized by a neural network \citep{VAE}. The goal is to train the model on data $X$ drawn from a data distribution $p_D(X)$ by minimizing the forward KL divergence $D_{KL}( p_D(X) \, || \, q_\theta(X)) = - \mathbb{E}_{X \sim p_D(X)} [ \log q_\theta(X)] + C$. As $\log q_\theta(X) = \log \left (\int q_\theta(X|Z) q(Z) \, dZ \right ) $  cannot be efficiently evaluated, the negative ELBO is estimated via an encoder network $p_{\Phi}(Z|X)$ and used as an upper bound on the forward KL divergence. Up to an additive constant, the ELBO objective is equivalent to the joint forward KL divergence $D_{KL}(p_{D, \Phi}(X,Z) \, || \, q_\theta(X,Z))$ \citep{kingma2019introduction}. In
\citet{ambrogioni2018forward} VAEs are used in variational inference by using KL divergences between joint-probability distributions. 

\textbf{Diffusion Models.}
Diffusion models \citep{DicksteinDiff, DenoisingDiffusionModels, SohlDiff} are latent variable models that are typically trained with samples $X_0$ from a data distribution $p_D(X_0)$. In the forward diffusion process noise is sequentially added at time step $t$ on $X_t$ by sampling a predefined noise distribution $p(X_t|X_{t-1})$. Starting from $X_0$ this process is repeated $T$ times until the resulting data follows approximately a known stationary distribution $q(X_T)$.
The diffusion model $q_\theta(X_{t-1}|X_t)$ learns the reverse diffusion process, i.e.~to iteratively remove the noise from samples of the stationary distribution. Once trained, the repeated application of this model to samples of the stationary distribution yields a generative process that approximates sampling from  $p_D(X_0)$. 
As in the case of VAEs, sample probabilities cannot be efficiently computed. Therefore, these models are typically trained via a loss that is given by the negative ELBO.
Discrete diffusion models are studied in \citet{DiscreteDiffusionModels}, where several different discrete noise distributions like the Categorical Noise Distribution used in this work are proposed.

\textbf{Neural Optimization.} 
Optimization using neural networks is a problem that is extensively researched. The objective is to find the optimal solution $\eta$ of an objective, i.e.~$\min_\eta \, O(\eta)$, where $\eta$ can either be either a function or a variable. Usually, the goal is to find $\eta$ without using training data. In UCO, for example, $\eta \in \{ 0,1\}^N$ is a vector with binary variables. As described in Sec.~\ref{sec:descr} this is frequently framed as a Neural Probabilistic Optimization (NPO) problem, where the goal is to approximate a target distribution with a neural network. \citet{wu_solving_2018} use autoregressive models to approximate discrete probability distributions related to Ising models. Another example of NPO are Boltzmann Generators \citep{BoltzmannGen}, where Normalizing Flows are used to learn the equilibrium distribution of molecule configurations at a given temperature. However, their method cannot be used in CO as Normalizing Flows are not applicable in the discrete setting. In \citet{NeuralImportanceSampling} a Normalizing Flow is used for Monte Carlo Integration. Another example of NPO is Variational Monte Carlo, where neural networks are used to approximate the lowest energy state of a quantum system \citep{VMC}.


\textbf{Approximate Likelihood Models in Neural Probabilistic Optimization}
In \citet{wu2020stochastic} Normalizing Flows are trained to approximate an annealed target distribution in continuous NPO problems. Here, MCMC is used to introduce randomness in the iterative sampling process and the data processing inequality is used to obtain an upper bound of the reverse KL divergence to train the model. In \citet{ContDiffModels4,ghio2023sampling, ContDiffModels1, ContDiffModels3, ContDiffModels2, akhound2024iterated} diffusion models are used in the continuous setting of NPO.

\textbf{Neural Combinatorial Optimization.}
Neural CO aims at the fast generation of high-quality solutions to CO problems. In supervised CO, a generative model is typically trained using solutions to CO problems that are obtained by classical solvers such as Gurobi \citep{gurobi} or KaMIS \citep{KaMIS}. In \citet{DIFUSCO}, for example, discrete diffusion models are applied by training them on data from solvers. In \citet{INTEL} (INTEL) and \citet{DGL} (DGL) supervised learning is combined with tree search refinement.
As highlighted in \citet{yehuda_its_2020}, these supervised approaches suffer from the problem of expensive data generation, leading to a growing interest in Unsupervised CO (UCO). In UCO, the aim is to train a model that learns how to solve CO problems without relying on labeled training data. These methods commonly rely on exact likelihood models, such as mean-field models \citep{karalias_erdos_2020, sun_annealed_2022, wang_unsupervised_2023}. These works highlight that the calculation of expectation values in the context of UCO is particularly convenient with mean-field models due to their assumption of statistical independence among the modeled random variables. As outlined in Sec.~\ref{sec:joint_diff-obj} our diffusion model approach also enjoys this feature. In \citet{ahn_learning_2020} (LwtD) a mean-field model is combined with Reinforcement Learning (RL) to iteratively refine solutions. \citet{VAG-CO} demonstrate that the statistical independence assumption in mean-field models restricts the performance of these approaches on particularly hard CO problems. They show that more expressive exact likelihood models like autoregressive models exhibit performance benefits. However, these models suffer from high memory requirements and long sampling time which slows down the training process. \citet{VAG-CO} (VAG-CO) argue that autoregressive methods should be combined with RL to reduce the memory requirements and speed up training. \citet{khalil_learning_2017-1} use autoregressive models combined with Q-learning to select the next solution variable together with its value. \citet{VAG-CO} address the problem of slow sampling in autoregressive models by introducing Subgraph Tokenization. \citet{gflow_2023} use GFlow networks \citep{Gflow_foundations}, which realizes autoregressive solution generation in UCO via GFlowNets \citep{bengio2021flow}. 

\begin{table*}[ht]
    \centering
    \begin{minipage}{0.50\textwidth}
    \setlength\tabcolsep{2pt}
    \begin{adjustbox}{width=\textwidth}
        \begin{tabular}{c c c c c c}
             \textbf{MIS} & & RB-small &  & RB-large & \\
              \hline
            \hline
            \rule{0pt}{12pt}
            Method & Type & Size $\uparrow$ & time $\downarrow$ & Size $\uparrow$ & time $\downarrow$\\
            \hline
            \rule{0pt}{12pt}
            Gurobi (r)& OR & 19.98 & 47:34 & 40.90 & 2:10:26 \\
            KaMIS (r)& OR & 20.10 & 1:24:12 & 43.15 & 2:03:36 \\
            \hline
            \rule{0pt}{12pt}
            LwtD (r)& UL & 19.01 & 1:17 & 32.32 & 7:33\\
            INTEL (r)& SL&  18.47 & 13:04 & 34.47 & 20:17\\
             DGL (r)& SL& 17.36 & 12:47 & 34.50 & 23:54\\
             LTFT (r)& UL& $\mathbf{19.18}$  & 0:32 & $37.48$ & 4:22\\
            \hline
            DiffUCO& UL& $18.88 \pm 0.06$ & 0:07 & $38.10 \pm 0.13$ & 0:10\\
            DiffUCO: CE& UL&  $\mathbf{19.24 \pm 0.05}$  & 0:54 & $\mathbf{38.87 \pm 0.13}$ & 4:57\\
            DiffUCO: CE-$\mathrm{ST}_{8}$ & UL& $ \mathbf{19.24 \pm 0.05 }$  & 00:13 & $\mathbf{38.87 \pm 0.13}$  & 1:05\\
    
        \end{tabular}
    \end{adjustbox}

    \end{minipage}
    \hfill
    \begin{minipage}{0.49\textwidth}
    \centering
    \setlength\tabcolsep{2pt}
    \begin{adjustbox}{width=\textwidth}
    \begin{tabular}{c c c c c c}
         \textbf{MDS} & & BA-small & & BA-large &\\
          \hline
        \hline
        \rule{0pt}{12pt}
        Method  & Type & Size $\downarrow$ & time $\downarrow$ & Size $\downarrow$ & time $\downarrow$\\
        \hline
        \rule{0pt}{12pt}
        Gurobi (r) & OR & 27.89 & 1:47 & 103.80 & 13:48\\
        \hline
        \rule{0pt}{12pt}
        Greedy (r) & H & 37.39 & 2:13 & 140.52 & 35:01\\
        MFA (r) & H & 36.36  & 2:56 & 126.56 & 36:31\\
         EGN: CE (r) &  UL &30.68 & 1:00 & 116.76 & 3:56\\
         EGN-Anneal: CE (r) & UL&  1:01 & 3:55  & 111.50 & 3:55\\
         LTFT (r) & UL & 28.61  & 2:20 & 110.28 & 32:12\\
        \hline
        \rule{0pt}{12pt}
        DiffUCO & UL & $28.30 \pm 0.10$ & 0:05 & $107.01 \pm 0.33$ & 0:05\\
        DiffUCO: CE & UL & $\mathbf{28.20 \pm 0.09}$ & 00:54 & $\mathbf{106.61 \pm 0.30}$ & 3:28\\
        DiffUCO: CE-$\mathrm{ST}_{8}$ & UL & $\mathbf{28.20 \pm 0.09}$  & 0:13 & $\mathbf{106.61 \pm 0.31}$  & 0:40\\

    \end{tabular}
    \end{adjustbox}
    \end{minipage}
    \vspace{-1ex}
    \caption{Left: Average independent set size on the whole test dataset on the RB-small and RB-large dataset. The higher the better. Right: Average dominating set size on the whole test dataset on the BA-small and BA-large datasets. The lower the set size the better. Left and Right: Total evaluation time is shown in h:m:s. (r) indicates that results are reported as in \citet{gflow_2023}. (CE) indicates that results are reported after applying Conditional Expectation and ($\mathrm{ST}_k$) indicates that results are reported by combining CE and ST at a specific $ k $ for faster CE. The best neural method is marked as bold.}
    \label{tab:MDS}
    \label{tab:MIS}
\end{table*}

\begin{table}[h]
\centering
\small
\setlength\tabcolsep{2pt}
\begin{adjustbox}{width=.5\textwidth}
{\renewcommand{\arraystretch}{1.8}
\begin{tabular}{c c}
  Problem Type  & Objective: $\mathop{\min}_{X \in \{ 0,1 \}^N} H(X)$  \\
\hline
\hline
  MIS &   
   $H(X) = - A \, \sum_{i= 1}^{N} X_i + B \, \sum_{(i,j) \in \mathcal{E}} X_i \cdot X_j$
   \\
\hline
 MDS &   
     $H(X) = A \, \sum_{i = 1}^N X_i + B \, \sum_{i = 1}^N (1-X_i) \prod_{j \in \mathcal{N}(j)} (1-X_j)$ \\
\hline
 MaxCl &   
     $H(X) = - A \, \sum_{i=1}^{N} X_i + B \, \sum_{(i,j) \notin \mathcal{E}} X_i \cdot X_j$
  \\
\hline
 MaxCut &   
         $H(\sigma) = - \sum_{(i,j) \in \mathcal{E}} \frac{1-\sigma_i \sigma_j}{2} \quad \mathrm{where} \, \, \sigma_i = 2 \, X_i -1$
         
 \\
 \hline
 MVC &  
    $H(X) = A \, \sum_{i=1}^{N} X_i + B \, \sum_{(i,j) \in \mathcal{E}} (1 - X_i) \cdot (1-X_j)$
       \\

\end{tabular}

}
\end{adjustbox}
\vspace{-3ex}
\caption{Table with energy functions of the MIS, MDS, MaxCl, MaxCut and MVC problems \citep{lucas_ising_2014}. Choosing $A < B$ ensures that all minima of the energy function are feasible solutions. In all of our Experiments, we chose $A = 1.0$ and $ B = 1.01$.}
\label{tab:COInstances}
\end{table}

\section{Experiments}
In this work, we evaluate our method DiffUCO on five different CO problem types, namely Maximum Independent Set (MIS), Maximum Clique (MaxCl), Minimum Dominating Set (MDS), Maximum Cut (MaxCut), and Minimum Vertex Cover (MVC). The energy functions used by DiffUCO for these CO problems can be found in Tab.~\ref{tab:COInstances}. To compare our method to recently published state-of-the-art autoregressive models, we compare on MIS, MaxCl, MDS, and MaxCut to published results reported in Let The Flows Tell (LTFT, \citet{gflow_2023}) (see Tab.~\ref{tab:MIS} and Tab.~\ref{tab:MaxCl}) and on MVC we compare our method to published results from \citet{VAG-CO} (see. Fig.~\ref{fig:MVC}, Left).
For the experiments on MIS and MaxCl, graphs are generated by the so-called RB-Model \citep{xu_hard_sat_instances_2005}, which is known to yield hard instances of CO problems like MIS, MVC and MaxCl \citep{toenshoff_graph_2020}. On MDS and MaxCut randomly generated Barabasi-Albert (BA) graphs \citep{albert_scaling_random_networks_1999} are used. For both of these graph types, 4000 graphs are used for training and 500 graphs are used for the validation and testing, respectively. In either case, we follow \citet{gflow_2023} and generate a small and a large dataset with uniformly random sampled graph sizes between 200-300 and 800-1200 nodes.
In accordance with \citet{gflow_2023} we label the methods in our tables either as Operations Research (OR), Supervised Learning (SL), or Unsupervised Learning (UL). On BA datasets, two Heuristic (H) baselines namely Greedy and mean-field annealing (MFA) are additionally used \citep{MFA}. The OR methods have substantially longer running times than the other methods, as they are usually run until the optimal solution is found. Therefore, these results cannot be used to compare the OR methods to the other methods. The OR results are merely provided to showcase the quality of optimal solutions to the CO problem datasets.
For each method we provide the inference time required to solve the test set problems (see App.~\ref{app:time}).\\
First, we report the performance of our method DiffUCO with and without CE and we investigate the runtime advantage of combining CE with ST (see Sec. \ref{sec:CE}). We denote the combination of CE and ST as DiffUCO: CE-ST$_{k}$, where $k$ is the token size used for ST. As DiffUCO: CE-ST$_{1}$ equates to CE without using ST, we will just denote it as DiffUCO: CE.
For the experiments in Tab.~\ref{tab:MIS} and Tab.~\ref{tab:MaxCl} we report our results as a test set average over eight samples per CO instance which is then averaged over three training seeds.
For more details about the experiments on MVC we refer to the corresponding paragraphs.
In Sec.~\ref{sec:ablations} we show that DiffUCO`s solution quality can further be improved when more diffusion steps are applied during evaluation time than we use during training. Therefore, on benchmark datasets, we always evaluate our method with three times more diffusion steps than we have used during training (see App.~\ref{app:evaluation_diff}).

\subsection{Benchmarks}
\label{sec:benchmarks}

\textbf{Maximum Independent Set.}
The MIS problem is the problem of finding the largest set within a graph under the constraint that neighboring nodes are not both within the set. We evaluate our method on RB-small and RB-large and compare to published results from \citet{gflow_2023}. Results are shown in Tab.~\ref{tab:MIS} (Left). The CE and CE-ST variants of DiffUCO achieve the best results on RB-large and marginally outperform LTFT on RB-small.

\textbf{Minimum Dominating Set.}
On the MDS problem, the goal is to find the set with the lowest number of vertices in a graph so that at least one neighbor of each node is within the set. We evaluate our method on the BA-small and BA-large datasets and compare in Tab.~\ref{tab:MDS} (Right) our results to published results of \citet{gflow_2023}. Here, DiffUCO and its variants outperform all other methods on both datasets. 

\textbf{Maximum Clique.}
In MaxCl the goal is to find the largest set of nodes, in which every node is connected to all other nodes within the set.
The MaxCl problem can be solved by solving the MIS problem on the complementary graph (see Tab.~\ref{tab:COInstances}). We evaluate our method on the RB-small datasets. Results are shown in Tab.~\ref{tab:MaxCl} (Left). DiffUCO CE-ST achieves the best results and outperforms LTFT insignificantly.

\textbf{Maximum Cut.}
The goal of the MaxCut problem is to separate a graph into two different sets of nodes so that the number of edges between these two sets is as high as possible. On MaxCut we evaluate our method on BA-small and BA-large graphs, where we see in Tab.~\ref{tab:MaxCut} (Right) that DiffUCO and its variants are the best-performing UCO methods. On BA-large our methods even outperform Gurobi with a long time limit of $300$ seconds per graph.

\begin{table*}[h]

    \centering
    \setlength\tabcolsep{2pt}
    \begin{adjustbox}{width=.85\textwidth}
    \begin{tabular}{c c c c c c c c c c}
         \textbf{MaxCl} & & RB-small & & \textbf{MaxCut} &  & BA-small & & BA-large &\\
          \hline
        \hline
        \rule{0pt}{12pt}
        Method & Type & Size $\uparrow$ & time $\downarrow$ & Method & Type & Size $\uparrow$ & time $\downarrow$ & Size $\uparrow$ & time $\downarrow$\\
        \hline
        \rule{0pt}{12pt}
        Gurobi (r) & OR & 19.05  & 1:55 & Gurobi $ t_{lim}$ & OR & $730.87 \pm 2.35$  & 8:30:00 & $2944.38 \pm 0.86$  & 1:17:35:00 \\
        \hline
        \rule{0pt}{12pt}
        Greedy (r) & H & 13.53  & 0:25 & Greedy (r) & H & 688.31 & 0:13 & 2786.00 & 3:07 \\
        MFA (r) & H &  14.82 & 0:27 & MFA (r) & H & 704.03 & 1:36 & 2833.86 & 7:16 \\
         EGN: CE (r) & UL & 12.02 & 0:41 & EGN: CE (r) & UL & 693.45 & 0:46 & 2870.34 & 2:49 \\
         EGN-Anneal: CE (r) & UL & 14.10  & 2:16 & EGN-Anneal: CE (r) & UL & 696.73  & 0:45 & 2863.23 & 2:48\\
         LTFT (r) &UL & $\mathbf{16.24}$  & 0:42 & LTFT (r) & UL & 704  & 2:57 & 2864 & 21:20 \\
        \hline
        DiffUCO & UL & $14.51 \pm 0.39$ & 0:04 & DiffUCO & UL& $ \mathbf{727.11 \pm 2.31}$ & 0:04 &  $\mathbf{2947.27 \pm 1.50} $& 0:04\\
        DiffUCO: CE & UL & $16.22 \pm 0.09$ & 1:00  & DiffUCO: CE & UL&  $ \mathbf{727.32 \pm 2.33}$ & 1:00 & $\mathbf{2947.53 \pm 1.48}$  & 3:47\\
        DiffUCO: CE-$\mathrm{ST}_{8}$ & UL & $\mathbf{16.30 \pm 0.08}$  & 0:13 
 & DiffUCO: CE-$\mathrm{ST}_{8}$ & UL & $ \mathbf{727.33 \pm 2.31}$  & 0:13  & $ \mathbf{2947.53 \pm 1.49}$ & 0:53\\

    \end{tabular}
    \end{adjustbox}
    \vspace{-1ex}
        \caption{Left: Testset average clique size on the whole RB-small dataset. The larger the set size the better. Right: Average test set cut size on the BA-small and BA-large datasets. The larger the better. Left and Right: Total evaluation time is shown in d:h:m:s. See Tab.~\ref{tab:MIS} for the meaning of (CE) and (CE-ST$_8$).  Gurobi $ t_{lim}$ denotes that Gurobi was run with a time limit. On BA-small the time limit is set to $60$ and on BA-large to $300$ seconds per graph. The best neural method is marked as bold.}
        \label{tab:MaxCut}
            \label{tab:MaxCl}
\end{table*}

\begin{figure*}[h]
    \centering
    \begin{minipage}{0.33\textwidth}
    \centering
    \setlength\tabcolsep{2pt}
  \includegraphics[width=1.\linewidth]{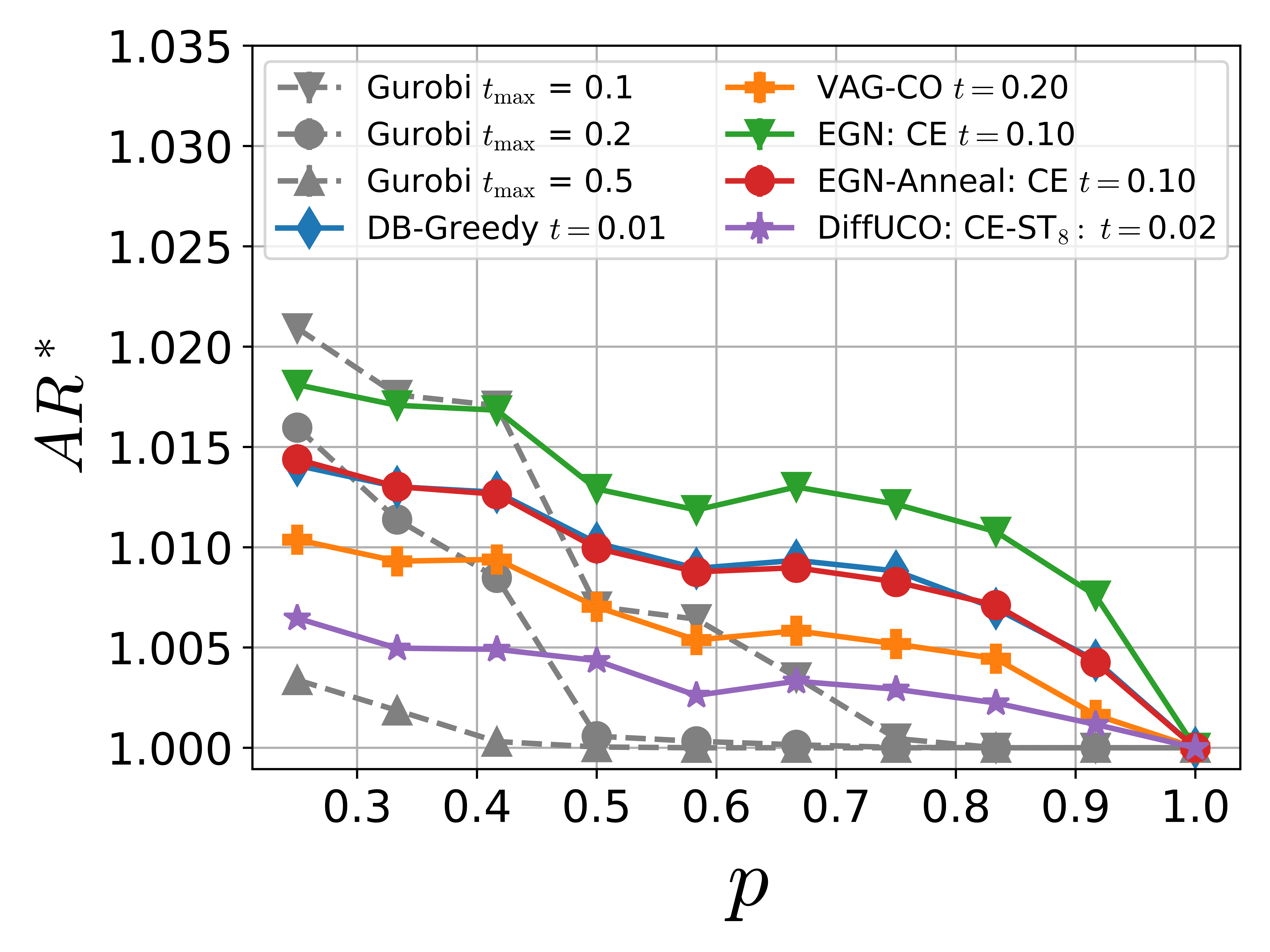}
  \end{minipage}
      \begin{minipage}{0.32\textwidth}
    \centering
    \setlength\tabcolsep{2pt}
  \includegraphics[width=1.\linewidth]{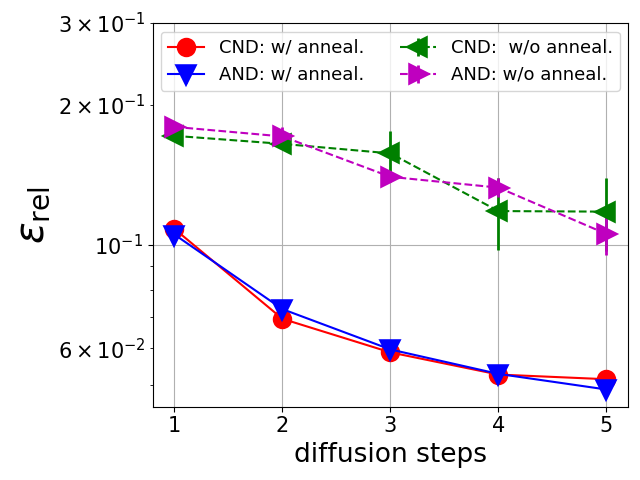}
  \end{minipage}
        \begin{minipage}{0.325\textwidth}
    \centering
    \setlength\tabcolsep{2pt}
  \includegraphics[width=.98\linewidth]{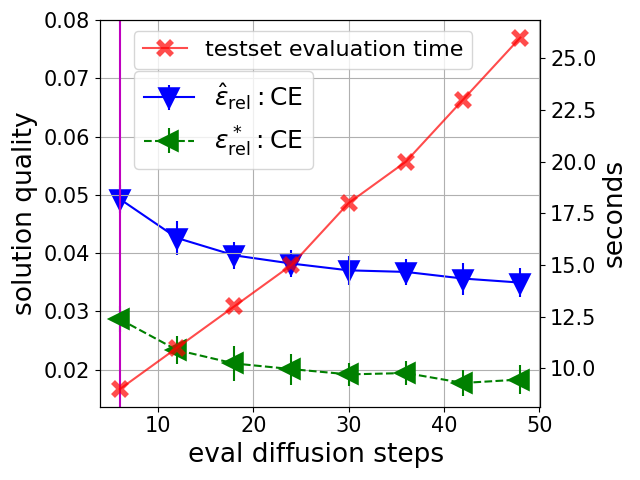}
  \end{minipage}
\vspace{-2ex}
\caption{Left: Comparison of $AR^*$ at different generation values $p$ on the RB-small MVC dataset. $t$ is the time that each method takes to generate the solutions. Middle: Relative error over the number of diffusion steps used during training for the Categorical Noise Distribution (CND) and the Annealed Noise Distributions (AND) in the setting with (w/ anneal) and without annealing (w/o anneal.). Right: Solution quality (left axis) on the RB-small MIS dataset over the number of diffusion steps during evaluation. $\hat{\epsilon}_\mathrm{rel}$ is the average relative error on the test set and $\epsilon_\mathrm{rel}^*$ is the best test set relative error out of $8$ model samples. The model is trained with $6$ diffusion steps (purple vertical line). }
\label{fig:MVC}
\end{figure*}

\textbf{Minimum Vertex Cover.}
The MVC problem aims at finding the smallest set of nodes so that every edge is connected to at least one node in the set.
On MVC we evaluate our method on the RB-200 dataset (see App.~\ref{app:datasets}) as reported in \citet{VAG-CO}, where the method VAG-CO is compared to EGN, EGN-Anneal, the greedy baseline DB-Greedy \citep{toenshoff_graph_2020} and Gurobi at different time limits. We compare to reported results from \citet{VAG-CO}, where the test set average of the best approximation ratio $AR^*$ (the lower the better, see App.~\ref{app:metrics}) is reported at different generation parameter values of $p$. The parameter $p$ controls the hardness of the CO problem on RB-graphs, where lower values of $p$ yield harder instances \citep{toenshoff_graph_2020}. Results are shown in Fig.~\ref{fig:MVC} (Left), where we see that DiffUCO significantly outperforms all other UCO methods on all values for $p$. On hard instances, i.e.~low values of $p$, DiffUCO achieves a significantly lower $AR^*$ than Gurobi, while needing only a fraction of its runtime. However, we note that a fair comparison of our method to Gurobi is difficult as Gurobi runs on CPUs and DiffUCO mostly on GPUs.

\textbf{Combination of CE and ST.}
Our experiments show that on MIS and MaxCl the use of CE improves the performance of our model significantly and that by using ST the computational cost of CE can be reduced by a large margin (see Tab.~\ref{tab:MIS} and Tab.~\ref{tab:MaxCl}).

\subsection{Ablations}
\label{sec:ablations}

Figure \ref{fig:MVC} (Middle) shows that increasing the number of diffusion steps during training consistently improves the performance of DiffUCO. For this, we train the model for different numbers of diffusion steps $T \in \{ 1,2,3,4,5 \} $ on the RB-100 MIS dataset (see App.~\ref{app:datasets}) and observe in Fig.~\ref{fig:MVC} that the relative error $\epsilon_{\mathrm{rel}}$ (the lower the better, see App.~\ref{app:metrics}) decreases as $T$ is increased, irrespectively of the used noise distribution. We also see that the Categorical Noise Distribution (CND) (blue) performs similarly to the Annealed Noise Distribution (AND) (orange) and that the use of annealing is highly beneficial. If we compare CND w/o annealing to AND w/o annealing we find that the results of the CND have a higher variance but similar average results. A possible reason for this is that the second term on the right-hand side of Eq.~\ref{eq:loss} is zero for the CND when no annealing is used. Therefore, this noise loss term provides no guidance at intermediate diffusion steps which may make the training process more prone to being stuck in different local minima depending on different weight initialization. We conclude from this that the AND might be the preferable noise distribution when no or only a little annealing is used.

\textbf{More evaluation diffusion steps improve solution quality.}
In Fig.~\ref{fig:MVC} (Right) we investigate whether the solution quality of DiffUCO can be improved by applying more diffusion steps during evaluation than during training. Here, the model is trained with $6$ diffusion steps (purple vertical line) on the RB-small MIS dataset. Figure \ref{fig:MVC} (Right) shows on the left axis the test dataset average relative error $\hat{\epsilon}_\mathrm{rel}$ and the best relative error $\epsilon_\mathrm{rel}^*$ out of $8$ solutions per problem instance over an increasing number of diffusion steps during evaluation (for details see App.~\ref{app:evaluation_diff}). We also plot on the additional y-axis the evaluation time on the whole dataset. The results show that both, $\hat{\epsilon}_\mathrm{rel}$ and $\epsilon_\mathrm{rel}^*$ consistently improve as the number of steps is increased. This improvement comes at the cost of a linearly increasing evaluation time cost.

\section{Limitations and Future Work}
While our method yields a model that can be used to generate high-quality solutions, the learned probability distribution will in practice always have a bias, i.e.~the model samples do not exactly follow the target distribution. In several interesting applications, a learned probability function is only useful when it allows to obtain unbiased samples from the target distribution \citep{UnbiClusterUpdates, unbiased1, unbiased2}.
Training DiffUCO on large graph datasets with high connectivity is memory- and time-expensive (see App.~\ref{app:resources}). Therefore, it would be interesting to consider approaches that improve upon these two aspects. This could be achieved by reducing the computational graph during backpropagation. The concept of latent diffusion models \citep{LatentDiff} and U-Net inspired GNN architectures \citep{Unet} represent promising research directions in this context.

\FloatBarrier

\section{Conclusion}
In this work, we introduce the Joint Variational Upper Bound as an efficiently computable loss that allows the application of latent variable models like diffusion models in Neural Probabilistic Optimization. Specifically, we demonstrate the excellent performance of diffusion models in Unsupervised Combinatorial Optimization. Our method outperforms recently presented methods on a wide range of benchmarks. 
Finally, we show in experiments that the application of variational annealing and additional diffusion steps during inference consistently improves the solution quality of the model.

\section*{Acknowledgements}
The ELLIS Unit Linz, the LIT AI Lab, the Institute for Machine Learning, are supported by the Federal State Upper
Austria. We thank the projects AI-MOTION (LIT-2018-
6-YOU-212), DeepFlood (LIT-2019-8-YOU-213), Medical Cognitive Computing Center (MC3), INCONTROLRL (FFG-881064), PRIMAL (FFG-873979), S3AI (FFG872172), DL for GranularFlow (FFG-871302), EPILEPSIA (FFG-892171), AIRI FG 9-N (FWF-36284, FWF36235), AI4GreenHeatingGrids(FFG- 899943), INTEGRATE (FFG-892418), ELISE (H2020-ICT-2019-3 ID:
951847), Stars4Waters (HORIZON-CL6-2021-CLIMATE01-01). We thank NXAI GmbH, Audi.JKU Deep Learning
Center, TGW LOGISTICS GROUP GMBH, Silicon Austria Labs (SAL), FILL Gesellschaft mbH, Anyline GmbH,
Google, ZF Friedrichshafen AG, Robert Bosch GmbH, UCB
Biopharma SRL, Merck Healthcare KGaA, Verbund AG,
GLS (Univ. Waterloo) Software Competence Center Hagenberg GmbH, TUV Austria, Frauscher Sensonic, TRUMPF ¨
and the NVIDIA Corporation. We acknowledge EuroHPC
Joint Undertaking for awarding us access to Karolina at
IT4Innovations.

\section*{Impact Statement}
In this work, we aim to improve data-free neural optimization algorithms. 
We hope our work will contribute to advances in various scientific fields such as Combinatorial Optimization.

\clearpage
\newpage
\bibliography{bibfile.bib}

\begin{thebibliography}{64}
\providecommand{\natexlab}[1]{#1}
\providecommand{\url}[1]{\texttt{#1}}
\expandafter\ifx\csname urlstyle\endcsname\relax
  \providecommand{\doi}[1]{doi: #1}\else
  \providecommand{\doi}{doi: \begingroup \urlstyle{rm}\Url}\fi

\bibitem[Ahn et~al.(2020)Ahn, Seo, and Shin]{ahn_learning_2020}
Sungsoo Ahn, Younggyo Seo, and Jinwoo Shin.
\newblock Learning what to defer for maximum independent sets.
\newblock In \emph{Proceedings of the 37th International Conference on Machine Learning, {ICML} 2020, 13-18 July 2020, Virtual Event}, volume 119 of \emph{Proceedings of Machine Learning Research}, pages 134--144. {PMLR}, 2020.
\newblock URL \url{http://proceedings.mlr.press/v119/ahn20a.html}.

\bibitem[Akhound-Sadegh et~al.(2024)Akhound-Sadegh, Rector-Brooks, Bose, Mittal, Lemos, Liu, Sendera, Ravanbakhsh, Gidel, Bengio, et~al.]{akhound2024iterated}
Tara Akhound-Sadegh, Jarrid Rector-Brooks, Avishek~Joey Bose, Sarthak Mittal, Pablo Lemos, Cheng-Hao Liu, Marcin Sendera, Siamak Ravanbakhsh, Gauthier Gidel, Yoshua Bengio, et~al.
\newblock Iterated denoising energy matching for sampling from boltzmann densities.
\newblock \emph{arXiv preprint arXiv:2402.06121}, 2024.

\bibitem[Ambrogioni et~al.(2018)Ambrogioni, G{\"{u}}{\c{c}}l{\"{u}}, Berezutskaya, van~den Borne, G{\"{u}}{\c{c}}l{\"{u}}t{\"{u}}rk, Hinne, Maris, and van Gerven]{ambrogioni2018forward}
Luca Ambrogioni, Umut G{\"{u}}{\c{c}}l{\"{u}}, Julia Berezutskaya, Eva W.~P. van~den Borne, Yagmur G{\"{u}}{\c{c}}l{\"{u}}t{\"{u}}rk, Max Hinne, Eric Maris, and Marcel A.~J. van Gerven.
\newblock Forward amortized inference for likelihood-free variational marginalization.
\newblock \emph{CoRR}, abs/1805.11542, 2018.
\newblock URL \url{http://arxiv.org/abs/1805.11542}.

\bibitem[Austin et~al.(2021)Austin, Johnson, Ho, Tarlow, and van~den Berg]{DiscreteDiffusionModels}
Jacob Austin, Daniel~D. Johnson, Jonathan Ho, Daniel Tarlow, and Rianne van~den Berg.
\newblock Structured denoising diffusion models in discrete state-spaces.
\newblock In \emph{Advances in Neural Information Processing Systems 34: Annual Conference on Neural Information Processing Systems 2021, NeurIPS 2021, December 6-14, 2021, virtual}, pages 17981--17993, 2021.
\newblock URL \url{https://proceedings.neurips.cc/paper/2021/hash/958c530554f78bcd8e97125b70e6973d-Abstract.html}.

\bibitem[Ba et~al.(2016)Ba, Kiros, and Hinton]{ba_layernorm_2016}
Lei~Jimmy Ba, Jamie~Ryan Kiros, and Geoffrey~E. Hinton.
\newblock Layer normalization.
\newblock \emph{CoRR}, abs/1607.06450, 2016.
\newblock URL \url{http://arxiv.org/abs/1607.06450}.

\bibitem[Barabási and Albert(1999)]{albert_scaling_random_networks_1999}
Albert-László Barabási and Réka Albert.
\newblock Emergence of scaling in random networks.
\newblock \emph{Science}, 286\penalty0 (5439):\penalty0 509--512, 1999.
\newblock \doi{10.1126/science.286.5439.509}.
\newblock URL \url{https://www.science.org/doi/abs/10.1126/science.286.5439.509}.

\bibitem[Barrett et~al.(2020)Barrett, Clements, Foerster, and Lvovsky]{ECO-DQN}
Thomas~D. Barrett, William~R. Clements, Jakob~N. Foerster, and A.~I. Lvovsky.
\newblock Exploratory combinatorial optimization with reinforcement learning.
\newblock In \emph{The Thirty-Fourth {AAAI} Conference on Artificial Intelligence, {AAAI} 2020, The Thirty-Second Innovative Applications of Artificial Intelligence Conference, {IAAI} 2020, The Tenth {AAAI} Symposium on Educational Advances in Artificial Intelligence, {EAAI} 2020, New York, NY, USA, February 7-12, 2020}, pages 3243--3250. {AAAI} Press, 2020.
\newblock \doi{10.1609/AAAI.V34I04.5723}.
\newblock URL \url{https://doi.org/10.1609/aaai.v34i04.5723}.

\bibitem[Barrett et~al.(2022)Barrett, Parsonson, and Laterre]{ECORD}
Thomas~D. Barrett, Christopher W.~F. Parsonson, and Alexandre Laterre.
\newblock Learning to solve combinatorial graph partitioning problems via efficient exploration.
\newblock \emph{CoRR}, abs/2205.14105, 2022.
\newblock \doi{10.48550/ARXIV.2205.14105}.
\newblock URL \url{https://doi.org/10.48550/arXiv.2205.14105}.

\bibitem[Bengio et~al.(2021{\natexlab{a}})Bengio, Jain, Korablyov, Precup, and Bengio]{bengio2021flow}
Emmanuel Bengio, Moksh Jain, Maksym Korablyov, Doina Precup, and Yoshua Bengio.
\newblock Flow network based generative models for non-iterative diverse candidate generation.
\newblock \emph{Advances in Neural Information Processing Systems}, 34:\penalty0 27381--27394, 2021{\natexlab{a}}.

\bibitem[Bengio et~al.(2021{\natexlab{b}})Bengio, Deleu, Hu, Lahlou, Tiwari, and Bengio]{Gflow_foundations}
Yoshua Bengio, Tristan Deleu, Edward~J. Hu, Salem Lahlou, Mo~Tiwari, and Emmanuel Bengio.
\newblock Gflownet foundations.
\newblock \emph{CoRR}, abs/2111.09266, 2021{\natexlab{b}}.
\newblock URL \url{https://arxiv.org/abs/2111.09266}.

\bibitem[Berner et~al.(2022)Berner, Richter, and Ullrich]{ContDiffModels4}
Julius Berner, Lorenz Richter, and Karen Ullrich.
\newblock An optimal control perspective on diffusion-based generative modeling.
\newblock \emph{arXiv preprint arXiv:2211.01364}, 2022.

\bibitem[Bilbro et~al.(1988)Bilbro, Mann, III, Snyder, van~den Bout, and White]{MFA}
Griff~L. Bilbro, Reinhold Mann, Thomas K.~Miller III, Wesley~E. Snyder, David~E. van~den Bout, and Mark~W. White.
\newblock Optimization by mean field annealing.
\newblock In \emph{Advances in Neural Information Processing Systems 1, {[NIPS} Conference, Denver, Colorado, USA, 1988]}, pages 91--98. Morgan Kaufmann, 1988.
\newblock URL \url{http://papers.nips.cc/paper/127-optimization-by-mean-field-annealing}.

\bibitem[B{\"{o}}ther et~al.(2022)B{\"{o}}ther, Ki{\ss}ig, Taraz, Cohen, Seidel, and Friedrich]{DGL}
Maximilian B{\"{o}}ther, Otto Ki{\ss}ig, Martin Taraz, Sarel Cohen, Karen Seidel, and Tobias Friedrich.
\newblock What's wrong with deep learning in tree search for combinatorial optimization.
\newblock In \emph{The Tenth International Conference on Learning Representations, {ICLR} 2022, Virtual Event, April 25-29, 2022}. OpenReview.net, 2022.
\newblock URL \url{https://openreview.net/forum?id=mk0HzdqY7i1}.

\bibitem[Bradbury et~al.(2018)Bradbury, Frostig, Hawkins, Johnson, Leary, Maclaurin, Necula, Paszke, Vander{P}las, Wanderman-{M}ilne, and Zhang]{jax2018github}
James Bradbury, Roy Frostig, Peter Hawkins, Matthew~James Johnson, Chris Leary, Dougal Maclaurin, George Necula, Adam Paszke, Jake Vander{P}las, Skye Wanderman-{M}ilne, and Qiao Zhang.
\newblock {JAX}: composable transformations of {P}ython+{N}um{P}y programs, 2018.
\newblock URL \url{http://github.com/google/jax}.

\bibitem[Cappart et~al.(2021)Cappart, Ch{\'{e}}telat, Khalil, Lodi, Morris, and Velickovic]{cappart_combinatorial_2022}
Quentin Cappart, Didier Ch{\'{e}}telat, Elias~B. Khalil, Andrea Lodi, Christopher Morris, and Petar Velickovic.
\newblock Combinatorial optimization and reasoning with graph neural networks.
\newblock In \emph{Proceedings of the Thirtieth International Joint Conference on Artificial Intelligence, {IJCAI} 2021, Virtual Event / Montreal, Canada, 19-27 August 2021}, pages 4348--4355. ijcai.org, 2021.
\newblock \doi{10.24963/ijcai.2021/595}.
\newblock URL \url{https://doi.org/10.24963/ijcai.2021/595}.

\bibitem[Carleo and Troyer(2017)]{VMC}
Giuseppe Carleo and Matthias Troyer.
\newblock Solving the quantum many-body problem with artificial neural networks.
\newblock \emph{Science}, 355\penalty0 (6325):\penalty0 602--606, 2017.
\newblock \doi{10.1126/science.aag2302}.
\newblock URL \url{https://www.science.org/doi/abs/10.1126/science.aag2302}.

\bibitem[Ghio et~al.(2023)Ghio, Dandi, Krzakala, and Zdeborov{\'a}]{ghio2023sampling}
Davide Ghio, Yatin Dandi, Florent Krzakala, and Lenka Zdeborov{\'a}.
\newblock Sampling with flows, diffusion and autoregressive neural networks: A spin-glass perspective.
\newblock \emph{arXiv preprint arXiv:2308.14085}, 2023.

\bibitem[Glover et~al.(2022)Glover, Kochenberger, Hennig, and Du]{QUBO}
Fred~W. Glover, Gary~A. Kochenberger, Rick Hennig, and Yu~Du.
\newblock Quantum bridge analytics {I:} a tutorial on formulating and using {QUBO} models.
\newblock \emph{Ann. Oper. Res.}, 314\penalty0 (1):\penalty0 141--183, 2022.
\newblock \doi{10.1007/S10479-022-04634-2}.
\newblock URL \url{https://doi.org/10.1007/s10479-022-04634-2}.

\bibitem[Goemans and Williamson(1995)]{SDP}
Michel~X. Goemans and David~P. Williamson.
\newblock Improved approximation algorithms for maximum cut and satisfiability problems using semidefinite programming.
\newblock \emph{J. {ACM}}, 42\penalty0 (6):\penalty0 1115--1145, 1995.
\newblock \doi{10.1145/227683.227684}.
\newblock URL \url{https://doi.org/10.1145/227683.227684}.

\bibitem[{Gurobi Optimization, LLC}(2023)]{gurobi}
{Gurobi Optimization, LLC}.
\newblock {Gurobi Optimizer Reference Manual}, 2023.
\newblock URL \url{https://www.gurobi.com}.

\bibitem[Hibat{-}Allah et~al.(2021)Hibat{-}Allah, Inack, Wiersema, Melko, and Carrasquilla]{hibat-allah_variational_2021}
Mohamed Hibat{-}Allah, Estelle~M. Inack, Roeland Wiersema, Roger~G. Melko, and Juan Carrasquilla.
\newblock Variational neural annealing.
\newblock \emph{Nat. Mach. Intell.}, 3\penalty0 (11):\penalty0 952--961, 2021.
\newblock \doi{10.1038/s42256-021-00401-3}.
\newblock URL \url{https://doi.org/10.1038/s42256-021-00401-3}.

\bibitem[Ho et~al.(2020)Ho, Jain, and Abbeel]{DenoisingDiffusionModels}
Jonathan Ho, Ajay Jain, and Pieter Abbeel.
\newblock Denoising diffusion probabilistic models.
\newblock In \emph{Advances in Neural Information Processing Systems 33: Annual Conference on Neural Information Processing Systems 2020, NeurIPS 2020, December 6-12, 2020, virtual}, 2020.
\newblock URL \url{https://proceedings.neurips.cc/paper/2020/hash/4c5bcfec8584af0d967f1ab10179ca4b-Abstract.html}.

\bibitem[Ji and Shen(2019)]{EUBO}
Chunlin Ji and Haige Shen.
\newblock Stochastic variational inference via upper bound.
\newblock \emph{CoRR}, abs/1912.00650, 2019.
\newblock URL \url{http://arxiv.org/abs/1912.00650}.

\bibitem[Karalias and Loukas(2020)]{karalias_erdos_2020}
Nikolaos Karalias and Andreas Loukas.
\newblock Erdos goes neural: an unsupervised learning framework for combinatorial optimization on graphs.
\newblock In \emph{Advances in Neural Information Processing Systems 33: Annual Conference on Neural Information Processing Systems 2020, NeurIPS 2020, December 6-12, 2020, virtual}, 2020.
\newblock URL \url{https://proceedings.neurips.cc/paper/2020/hash/49f85a9ed090b20c8bed85a5923c669f-Abstract.html}.

\bibitem[Khalil et~al.(2017)Khalil, Dai, Zhang, Dilkina, and Song]{khalil_learning_2017-1}
Elias~B. Khalil, Hanjun Dai, Yuyu Zhang, Bistra Dilkina, and Le~Song.
\newblock Learning combinatorial optimization algorithms over graphs.
\newblock In \emph{Advances in Neural Information Processing Systems 30: Annual Conference on Neural Information Processing Systems 2017, December 4-9, 2017, Long Beach, CA, {USA}}, pages 6348--6358, 2017.
\newblock URL \url{https://proceedings.neurips.cc/paper/2017/hash/d9896106ca98d3d05b8cbdf4fd8b13a1-Abstract.html}.

\bibitem[Kingma and Welling(2014)]{VAE}
Diederik~P. Kingma and Max Welling.
\newblock Auto-encoding variational bayes.
\newblock In \emph{2nd International Conference on Learning Representations, {ICLR} 2014, Banff, AB, Canada, April 14-16, 2014, Conference Track Proceedings}, 2014.
\newblock URL \url{http://arxiv.org/abs/1312.6114}.

\bibitem[Kingma et~al.(2019)Kingma, Welling, et~al.]{kingma2019introduction}
Diederik~P Kingma, Max Welling, et~al.
\newblock An introduction to variational autoencoders.
\newblock \emph{Foundations and Trends{\textregistered} in Machine Learning}, 12\penalty0 (4):\penalty0 307--392, 2019.

\bibitem[Kirkpatrick et~al.(1983)Kirkpatrick, Gelatt, and Vecchi]{sim_ann_1983}
S.~Kirkpatrick, C.~D. Gelatt, and M.~P. Vecchi.
\newblock Optimization by simulated annealing.
\newblock \emph{Science}, 220\penalty0 (4598):\penalty0 671--680, 1983.
\newblock \doi{10.1126/science.220.4598.671}.
\newblock URL \url{https://www.science.org/doi/abs/10.1126/science.220.4598.671}.

\bibitem[Lamm et~al.(2017)Lamm, Sanders, Schulz, Strash, and Werneck]{KaMIS}
Sebastian Lamm, Peter Sanders, Christian Schulz, Darren Strash, and Renato~F. Werneck.
\newblock Finding near-optimal independent sets at scale.
\newblock \emph{J. Heuristics}, 23\penalty0 (4):\penalty0 207--229, 2017.
\newblock \doi{10.1007/S10732-017-9337-X}.
\newblock URL \url{https://doi.org/10.1007/s10732-017-9337-x}.

\bibitem[Le and Jegelka(2023)]{jegelka1}
Thien Le and Stefanie Jegelka.
\newblock Limits, approximation and size transferability for gnns on sparse graphs via graphops.
\newblock In \emph{Advances in Neural Information Processing Systems 36: Annual Conference on Neural Information Processing Systems 2023, NeurIPS 2023, New Orleans, LA, USA, December 10 - 16, 2023}, 2023.
\newblock URL \url{http://papers.nips.cc/paper\_files/paper/2023/hash/8154c89c8d3612d39fd1ed6a20f4bab1-Abstract-Conference.html}.

\bibitem[Li et~al.(2018)Li, Chen, and Koltun]{INTEL}
Zhuwen Li, Qifeng Chen, and Vladlen Koltun.
\newblock Combinatorial optimization with graph convolutional networks and guided tree search.
\newblock In \emph{Advances in Neural Information Processing Systems 31: Annual Conference on Neural Information Processing Systems 2018, NeurIPS 2018, December 3-8, 2018, Montr{\'{e}}al, Canada}, pages 537--546, 2018.
\newblock URL \url{https://proceedings.neurips.cc/paper/2018/hash/8d3bba7425e7c98c50f52ca1b52d3735-Abstract.html}.

\bibitem[Lin et~al.(2021)Lin, Jaech, Li, Gormley, and Eisner]{lin2021limitations}
Chu-Cheng Lin, Aaron Jaech, Xin Li, Matthew~R Gormley, and Jason Eisner.
\newblock Limitations of autoregressive models and their alternatives.
\newblock In \emph{Proceedings of the 2021 Conference of the North American Chapter of the Association for Computational Linguistics: Human Language Technologies}, pages 5147--5173, 2021.

\bibitem[Liu et~al.(2020)Liu, Jiang, He, Chen, Liu, Gao, and Han]{RADAM}
Liyuan Liu, Haoming Jiang, Pengcheng He, Weizhu Chen, Xiaodong Liu, Jianfeng Gao, and Jiawei Han.
\newblock On the variance of the adaptive learning rate and beyond.
\newblock In \emph{8th International Conference on Learning Representations, {ICLR} 2020, Addis Ababa, Ethiopia, April 26-30, 2020}. OpenReview.net, 2020.
\newblock URL \url{https://openreview.net/forum?id=rkgz2aEKDr}.

\bibitem[Lucas(2014)]{lucas_ising_2014}
Andrew Lucas.
\newblock Ising formulations of many np problems.
\newblock \emph{Frontiers in Physics}, 2, 2014.
\newblock ISSN 2296-424X.
\newblock \doi{10.3389/fphy.2014.00005}.
\newblock URL \url{https://www.frontiersin.org/articles/10.3389/fphy.2014.00005}.

\bibitem[McNaughton et~al.(2020)McNaughton, Milo\ifmmode \check{s}\else \v{s}\fi{}evi\ifmmode~\acute{c}\else \'{c}\fi{}, Perali, and Pilati]{unbiased2}
B.~McNaughton, M.~V. Milo\ifmmode \check{s}\else \v{s}\fi{}evi\ifmmode~\acute{c}\else \'{c}\fi{}, A.~Perali, and S.~Pilati.
\newblock Boosting monte carlo simulations of spin glasses using autoregressive neural networks.
\newblock \emph{Phys. Rev. E}, 101:\penalty0 053312, May 2020.
\newblock \doi{10.1103/PhysRevE.101.053312}.
\newblock URL \url{https://link.aps.org/doi/10.1103/PhysRevE.101.053312}.

\bibitem[Minka et~al.(2005)]{minka2005divergence}
Tom Minka et~al.
\newblock Divergence measures and message passing.
\newblock Technical report, Technical report, Microsoft Research, 2005.

\bibitem[M{\"{u}}ller et~al.(2019)M{\"{u}}ller, McWilliams, Rousselle, Gross, and Nov{\'{a}}k]{NeuralImportanceSampling}
Thomas M{\"{u}}ller, Brian McWilliams, Fabrice Rousselle, Markus Gross, and Jan Nov{\'{a}}k.
\newblock Neural importance sampling.
\newblock \emph{{ACM} Trans. Graph.}, 38\penalty0 (5):\penalty0 145:1--145:19, 2019.
\newblock \doi{10.1145/3341156}.
\newblock URL \url{https://doi.org/10.1145/3341156}.

\bibitem[Nicoli et~al.(2020)Nicoli, Nakajima, Strodthoff, Samek, M\"uller, and Kessel]{unbiased1}
Kim~A. Nicoli, Shinichi Nakajima, Nils Strodthoff, Wojciech Samek, Klaus-Robert M\"uller, and Pan Kessel.
\newblock Asymptotically unbiased estimation of physical observables with neural samplers.
\newblock \emph{Phys. Rev. E}, 101:\penalty0 023304, Feb 2020.
\newblock \doi{10.1103/PhysRevE.101.023304}.
\newblock URL \url{https://link.aps.org/doi/10.1103/PhysRevE.101.023304}.

\bibitem[No{\'{e}} and Wu(2018)]{BoltzmannGen}
Frank No{\'{e}} and Hao Wu.
\newblock Boltzmann generators - sampling equilibrium states of many-body systems with deep learning.
\newblock \emph{CoRR}, abs/1812.01729, 2018.
\newblock URL \url{http://arxiv.org/abs/1812.01729}.

\bibitem[Raghavan(1988)]{raghavan1988probabilistic}
Prabhakar Raghavan.
\newblock Probabilistic construction of deterministic algorithms: approximating packing integer programs.
\newblock \emph{Journal of Computer and System Sciences}, 37\penalty0 (2):\penalty0 130--143, 1988.

\bibitem[Richter et~al.(2023)Richter, Berner, and Liu]{ContDiffModels1}
Lorenz Richter, Julius Berner, and Guan-Horng Liu.
\newblock Improved sampling via learned diffusions.
\newblock \emph{arXiv preprint arXiv:2307.01198}, 2023.

\bibitem[Rombach et~al.(2022)Rombach, Blattmann, Lorenz, Esser, and Ommer]{LatentDiff}
Robin Rombach, Andreas Blattmann, Dominik Lorenz, Patrick Esser, and Bj{\"{o}}rn Ommer.
\newblock High-resolution image synthesis with latent diffusion models.
\newblock In \emph{{IEEE/CVF} Conference on Computer Vision and Pattern Recognition, {CVPR} 2022, New Orleans, LA, USA, June 18-24, 2022}, pages 10674--10685. {IEEE}, 2022.
\newblock \doi{10.1109/CVPR52688.2022.01042}.
\newblock URL \url{https://doi.org/10.1109/CVPR52688.2022.01042}.

\bibitem[Ronneberger et~al.(2015)Ronneberger, Fischer, and Brox]{Unet}
Olaf Ronneberger, Philipp Fischer, and Thomas Brox.
\newblock U-net: Convolutional networks for biomedical image segmentation.
\newblock \emph{CoRR}, abs/1505.04597, 2015.
\newblock URL \url{http://arxiv.org/abs/1505.04597}.

\bibitem[Sanokowski et~al.(2022)Sanokowski, Berghammer, Kofler, Hochreiter, and Lehner]{sanokowski_one_2022}
Sebastian Sanokowski, Wilhelm Berghammer, Johannes Kofler, Sepp Hochreiter, and Sebastian Lehner.
\newblock One {Network} to {Approximate} {Them} {All}: {Amortized} {Variational} {Inference} of {Ising} {Ground} {States}.
\newblock \emph{Machine Learning and the Physical Sciences workshop, NeurIPS 2022}, 2022.

\bibitem[Sanokowski et~al.(2023)Sanokowski, Berghammer, Hochreiter, and Lehner]{VAG-CO}
Sebastian Sanokowski, Wilhelm Berghammer, Sepp Hochreiter, and Sebastian Lehner.
\newblock Variational annealing on graphs for combinatorial optimization.
\newblock In \emph{Advances in Neural Information Processing Systems 36: Annual Conference on Neural Information Processing Systems 2023, NeurIPS 2023, New Orleans, LA, USA, December 10 - 16, 2023}, 2023.
\newblock URL \url{http://papers.nips.cc/paper\_files/paper/2023/hash/c9c54ac0dd5e942b99b2b51c297544fd-Abstract-Conference.html}.

\bibitem[Sohl{-}Dickstein et~al.(2015)Sohl{-}Dickstein, Weiss, Maheswaranathan, and Ganguli]{DicksteinDiff}
Jascha Sohl{-}Dickstein, Eric~A. Weiss, Niru Maheswaranathan, and Surya Ganguli.
\newblock Deep unsupervised learning using nonequilibrium thermodynamics.
\newblock In \emph{Proceedings of the 32nd International Conference on Machine Learning, {ICML} 2015, Lille, France, 6-11 July 2015}, volume~37 of \emph{{JMLR} Workshop and Conference Proceedings}, pages 2256--2265. JMLR.org, 2015.
\newblock URL \url{http://proceedings.mlr.press/v37/sohl-dickstein15.html}.

\bibitem[Song et~al.(2021)Song, Sohl{-}Dickstein, Kingma, Kumar, Ermon, and Poole]{SohlDiff}
Yang Song, Jascha Sohl{-}Dickstein, Diederik~P. Kingma, Abhishek Kumar, Stefano Ermon, and Ben Poole.
\newblock Score-based generative modeling through stochastic differential equations.
\newblock In \emph{9th International Conference on Learning Representations, {ICLR} 2021, Virtual Event, Austria, May 3-7, 2021}. OpenReview.net, 2021.
\newblock URL \url{https://openreview.net/forum?id=PxTIG12RRHS}.

\bibitem[Sun et~al.(2022)Sun, Guha, and Dai]{sun_annealed_2022}
Haoran Sun, Etash~Kumar Guha, and Hanjun Dai.
\newblock Annealed training for combinatorial optimization on graphs.
\newblock In \emph{OPT 2022: Optimization for Machine Learning (NeurIPS 2022 Workshop)}, 2022.
\newblock URL \url{https://openreview.net/forum?id=fo3b0XjTkU}.

\bibitem[Sun and Yang(2023)]{DIFUSCO}
Zhiqing Sun and Yiming Yang.
\newblock {DIFUSCO:} graph-based diffusion solvers for combinatorial optimization.
\newblock \emph{CoRR}, abs/2302.08224, 2023.
\newblock \doi{10.48550/ARXIV.2302.08224}.
\newblock URL \url{https://doi.org/10.48550/arXiv.2302.08224}.

\bibitem[T{\"{o}}nshoff et~al.(2020)T{\"{o}}nshoff, Ritzert, Wolf, and Grohe]{toenshoff_graph_2020}
Jan T{\"{o}}nshoff, Martin Ritzert, Hinrikus Wolf, and Martin Grohe.
\newblock Graph neural networks for maximum constraint satisfaction.
\newblock \emph{Frontiers Artif. Intell.}, 3:\penalty0 580607, 2020.
\newblock \doi{10.3389/frai.2020.580607}.
\newblock URL \url{https://doi.org/10.3389/frai.2020.580607}.

\bibitem[T{\"{o}}nshoff et~al.(2023)T{\"{o}}nshoff, Kisin, Lindner, and Grohe]{anycsp}
Jan T{\"{o}}nshoff, Berke Kisin, Jakob Lindner, and Martin Grohe.
\newblock One model, any {CSP:} graph neural networks as fast global search heuristics for constraint satisfaction.
\newblock In \emph{Proceedings of the Thirty-Second International Joint Conference on Artificial Intelligence, {IJCAI} 2023, 19th-25th August 2023, Macao, SAR, China}, pages 4280--4288. ijcai.org, 2023.
\newblock \doi{10.24963/IJCAI.2023/476}.
\newblock URL \url{https://doi.org/10.24963/ijcai.2023/476}.

\bibitem[van Erven and Harremo{\"{e}}s(2014)]{RenyiDivergence}
Tim van Erven and Peter Harremo{\"{e}}s.
\newblock R{\'{e}}nyi divergence and kullback-leibler divergence.
\newblock \emph{{IEEE} Trans. Inf. Theory}, 60\penalty0 (7):\penalty0 3797--3820, 2014.
\newblock \doi{10.1109/TIT.2014.2320500}.
\newblock URL \url{https://doi.org/10.1109/TIT.2014.2320500}.

\bibitem[Vargas et~al.(2023)Vargas, Grathwohl, and Doucet]{ContDiffModels3}
Francisco Vargas, Will Grathwohl, and Arnaud Doucet.
\newblock Denoising diffusion samplers.
\newblock \emph{arXiv preprint arXiv:2302.13834}, 2023.

\bibitem[Vargas et~al.(2024)Vargas, Padhy, Blessing, and N{\"u}sken]{ContDiffModels2}
Francisco Vargas, Shreyas Padhy, Denis Blessing, and N~N{\"u}sken.
\newblock Transport meets variational inference: Controlled monte carlo diffusions.
\newblock In \emph{The Twelfth International Conference on Learning Representations}, 2024.

\bibitem[Wang and Li(2023)]{wang_unsupervised_2023}
Haoyu~Peter Wang and Pan Li.
\newblock Unsupervised learning for combinatorial optimization needs meta learning.
\newblock In \emph{The Eleventh International Conference on Learning Representations}, 2023.
\newblock URL \url{https://openreview.net/forum?id=-ENYHCE8zBp}.

\bibitem[Williams(1992)]{REINFORCE}
Ronald~J. Williams.
\newblock Simple statistical gradient-following algorithms for connectionist reinforcement learning.
\newblock \emph{Mach. Learn.}, 8:\penalty0 229--256, 1992.
\newblock \doi{10.1007/BF00992696}.
\newblock URL \url{https://doi.org/10.1007/BF00992696}.

\bibitem[Wu et~al.(2019)Wu, Wang, and Zhang]{wu_solving_2018}
Dian Wu, Lei Wang, and Pan Zhang.
\newblock Solving statistical mechanics using variational autoregressive networks.
\newblock \emph{Phys. Rev. Lett.}, 122:\penalty0 080602, Feb 2019.
\newblock \doi{10.1103/PhysRevLett.122.080602}.
\newblock URL \url{https://link.aps.org/doi/10.1103/PhysRevLett.122.080602}.

\bibitem[Wu et~al.(2021)Wu, Rossi, and Carleo]{UnbiClusterUpdates}
Dian Wu, Riccardo Rossi, and Giuseppe Carleo.
\newblock Unbiased monte carlo cluster updates with autoregressive neural networks.
\newblock \emph{Physical Review Research}, 3\penalty0 (4):\penalty0 L042024, 2021.

\bibitem[Wu et~al.(2020)Wu, K{\"o}hler, and No{\'e}]{wu2020stochastic}
Hao Wu, Jonas K{\"o}hler, and Frank No{\'e}.
\newblock Stochastic normalizing flows.
\newblock \emph{Advances in Neural Information Processing Systems}, 33:\penalty0 5933--5944, 2020.

\bibitem[Xu et~al.(2005)Xu, Boussemart, Hemery, and Lecoutre]{xu_hard_sat_instances_2005}
Ke~Xu, Fr{\'{e}}d{\'{e}}ric Boussemart, Fred Hemery, and Christophe Lecoutre.
\newblock A simple model to generate hard satisfiable instances.
\newblock In \emph{IJCAI-05, Proceedings of the Nineteenth International Joint Conference on Artificial Intelligence, Edinburgh, Scotland, UK, July 30 - August 5, 2005}, pages 337--342. Professional Book Center, 2005.
\newblock URL \url{http://ijcai.org/Proceedings/05/Papers/0989.pdf}.

\bibitem[Ye(2003)]{Gset}
Y.~Ye.
\newblock Gset.
\newblock 2003.
\newblock URL \url{https://web.stanford.edu/~yyye/yyye/Gset/}.

\bibitem[Yehuda et~al.(2020)Yehuda, Gabel, and Schuster]{yehuda_its_2020}
Gal Yehuda, Moshe Gabel, and Assaf Schuster.
\newblock It's not what machines can learn, it's what we cannot teach.
\newblock In \emph{Proceedings of the 37th International Conference on Machine Learning, {ICML} 2020, 13-18 July 2020, Virtual Event}, volume 119 of \emph{Proceedings of Machine Learning Research}, pages 10831--10841. {PMLR}, 2020.
\newblock URL \url{http://proceedings.mlr.press/v119/yehuda20a.html}.

\bibitem[Zhang et~al.(2023)Zhang, Dai, Malkin, Courville, Bengio, and Pan]{gflow_2023}
Dinghuai Zhang, Hanjun Dai, Nikolay Malkin, Aaron~C. Courville, Yoshua Bengio, and Ling Pan.
\newblock Let the flows tell: Solving graph combinatorial problems with gflownets.
\newblock In \emph{Advances in Neural Information Processing Systems 36: Annual Conference on Neural Information Processing Systems 2023, NeurIPS 2023, New Orleans, LA, USA, December 10 - 16, 2023}, 2023.
\newblock URL \url{http://papers.nips.cc/paper\_files/paper/2023/hash/27571b74d6cd650b8eb6cf1837953ae8-Abstract-Conference.html}.

\bibitem[Zhang et~al.(2021)Zhang, Li, Xia, Wang, and Jin]{jegelka2}
Muhan Zhang, Pan Li, Yinglong Xia, Kai Wang, and Long Jin.
\newblock Labeling trick: {A} theory of using graph neural networks for multi-node representation learning.
\newblock In \emph{Advances in Neural Information Processing Systems 34: Annual Conference on Neural Information Processing Systems 2021, NeurIPS 2021, December 6-14, 2021, virtual}, pages 9061--9073, 2021.
\newblock URL \url{https://proceedings.neurips.cc/paper/2021/hash/4be49c79f233b4f4070794825c323733-Abstract.html}.

\end{thebibliography}
\bibliographystyle{plainnat}
\newpage

\appendix
\onecolumn
\section{Derivations}
\subsection{Joint Variational Upper Bound}
\label{app:upper_bound}

Consider the simple case of a one-step diffusion model.
Here in the forward process the distribution $p_B(x)$ is directly mapped to the stationary distribution $q(z)$ by the application of a noise distribution $p(z|x)$.
In the reverse process, the parameterized distribution $q_\theta(x|z)$ directly maps samples from the stationary distribution $q(z)$ to the target distribution $p_B(x)$.

In the following we will derive that $ D_{KL}(q_\theta(x) \, || \, p_B(x))$ is upper bounded by $ D_{KL}(q_\theta(x,z) \, || \, p_B(x,z))$, where $p_B(x,z) = p(z|x) \, p_B(x)$ and $q_\theta(x,z) = q_\theta(x|z) \, q(z)$.

\textbf{Derivation via the chain-rule for Kl divergences:}\\
This inequality follows directly from the chain-rule for KL divergences which states that:

\begin{align*}
    D_{KL}(q_\theta(x,z) \, || \, p_B(x,z)) &= \int q_\theta(x,y) \log{\frac{q_\theta(x,y)}{p(x,y)}} d x \, d y \\
    &= \int q_\theta(x,z) \left [ \log{\frac{q_\theta(x)}{p(x)}} + \log{\frac{q_\theta(z|x)}{p(z|x)}} \right ] d x \, d z \\
    &= D_{KL}(q_\theta(x) \, || \, p_B(x)) + \mathbb{E}_{q_\theta(x) } \left [ D_{KL}(q_\theta(z|x) \, || \, p(z|x))\right ].
\end{align*}

According to Gibbs' inequality, the second term on the right-hand side is non-negative which is sufficient to show the upper bound.

\textbf{Derivation via the Evidence Upper Bound:}\\
Alternatively, the upper bound can be derived using the Evidence Upper Bound (EUBO) \citep{EUBO}, where we use $q_\theta(x) = \frac{q_\theta(x,z)}{q_\theta(z|x)}$ and $\int q_\theta(z|x) dz = 1 $ and therefore $$\log q_\theta(x)  = \int q_\theta(z | x)  \log{ q_\theta(x)}  dz = \int q_\theta(z | x) [ \log{ q_\theta(x,z)} - \log{ q_\theta(z|x)} ] dz$$

In the first step, we will derive an upper bound for  $\log{q_\theta(x) } $ by using the EUBO that is based on the Gibbs' inequality.

\begin{align*}
    \log q_\theta(x)  = \int q_\theta(z | x) [ \log{ q_\theta(x,z)} - \log{ q_\theta(z|x)} ] dz \leq \int q_\theta(z | x) [ \log{ q_\theta(x,z)} - \log{ p(z|x)} ] dz 
\end{align*}

By using this inequality we can now show that $ D_{KL}(q_\theta(x) \, || \, p_B(x)) \leq D_{KL}(q_\theta(x,z) \, || \, p_B(x,z))$.

\begin{align*}
    D_{KL}(q_\theta(x) \, || \, p_B(x)) &= \int q_\theta(x)  \log{ \frac{q_\theta(x) }{p_B(x)} }dx \leq  \int q_\theta(x) \left [ \int  q_\theta(z | x) \biggl ( \log{ q_\theta(x,z)} - \log{ p(z|x)} \biggr ) dz - \log{p_B(x)} \right ] dx \\
\end{align*}

By using $$ \int q_\theta(x) \log{ p_B(x)} dx =  \int \int q_\theta(x,z) \log{ p_B(x)} dz dx,$$ we can then write with $- \log{p(z|x)} - \log{p_B(x)} = - \log p_B(x,z)$

\begin{align*}
    D_{KL}(q_\theta(x) \, || \, p_B(x)) & \leq \int \int q_\theta(x,z) \left [ \log{q_\theta(x,z)} - \log{p(z|x)} - \log{p_B(x)} \right ] dz dx \\
    & = D_{KL}(q_\theta(x,z) \, || \, p_B(x,z)),
\end{align*}
 which proves the inequality.

\subsection{Joint Variational Diffusion Model Objective}
\label{app:deriv_obj}

In this section, we aim to derive the objective shown in Eq.\ref{eq:loss}.

Starting from
\begin{align*}
    D_{KL}(q_\theta (X_{0:T}) \, || \, p(X_{0:T}))  & = \sum_{X_{0:T} } q_\theta(X_{0:T}) \biggl [  \log{q_\theta(X_{0:T})} - \log{p(X_{0:T})} \biggr ] \\
    & = \sum_{X_{0:T} } q_\theta(X_{0:T}) \biggl [ \sum_{t= 1}^{T} \biggl [ \log{q_\theta(X_{t-1}|X_t)} - \log{p(X_{t}|X_{t-1})} \biggr ] + \log{q(X_T)} - \log{p_B(X_0)} \biggr ] 
\end{align*}

we can first simplify the first term with $$\sum_{X_{T:0}} q_\theta(X_{T:0}) \log{q_\theta(X_{t-1}|X_t)} = \sum_{X_{T:t-1}} q_\theta(X_{T:t-1}) \log{q_\theta(X_{t-1}|X_t)}$$ where we have used that the sum over all $X_\tau$ where $\tau < t-1$ sums up to one because $\log{q_\theta(X_{t-1}|X_t)}$ is independent of samples coming from these time steps. Remember that in the reverse process the sample process starts at $X_T$ and ends at $X_0$. So samples, where $\Tau < t - 1$ are future events and therefore $\log{q_\theta(X_{t-1}|X_t)}$ does not depend on these future events.
In the second step we pull out the sum over $X_{t-1}$, so that we arrive at

\begin{align*}
    \sum_{X_{T:t-1}} q_\theta(X_{T:t-1}) \log{q_\theta(X_{t-1}|X_t)} &= \sum_{X_{T:t}} q_\theta(X_{T:t}) \sum_{X_{t-1}} q_\theta(X_{t-1}|X_t) \log{q_\theta(X_{t-1}|X_t)} \\
    &= - \sum_{X_{T:t}} q_\theta(X_{T:t}) S(q_\theta(X_{t-1}|X_t)), 
\end{align*}

where $S(q_\theta(X_{t-1}|X_t))$ is the entropy over $q_\theta(X_{t-1}|X_t)$. 
Analogously, this simplification can be done with $\Tilde{H}(X_{t},X_{t-1})  := - \log{p(X_{t}|X_{t-1})}$, so that we arrive at:

\begin{align*}
    \sum_{X_{T:0}} q_\theta(X_{T:0}) \Tilde{H}(X_t,X_{t-1}) &= \sum_{X_{T:t-1}} q_\theta(X_{T:t-1}) \Tilde{H}(X_t,X_{t-1}) 
\end{align*}

Furthermore, we use that $\log{p_B(X_0)} = - \beta H(X_0) - \log{\mathcal{Z}}$ and absorb $\log(\mathcal{Z})$ and $\log{q(X_T)}$ into the constant $C$ as these terms neither depend on $\theta$ nor on $X_{0:T-1}$.
Finally, we arrive at:

\begin{align*}
    D_{KL}(q_\theta (X_{0:T}) \, || \, p(X_{0:T}))  & = 
     \sum_{t= 1}^{T} \sum_{X_{T:t} } q_\theta(X_{T:t}) \biggl [ \biggl [ S(q_\theta(X_{t-1}|X_t)) - \log{p(X_{t}|X_{t-1})} \biggr ] + \log{q(X_t)} - \log{p_B(X_0)} \biggr ] \\
     & = -  \sum_{t = 1}^{T} \mathbb{E}_{X_{T:t} \sim q_\theta(X_{T:t})} \left [ S(q_\theta(X_{t-1}|X_t)) \right ]  \\ & + \sum_{t=1}^{T} \mathbb{E}_{X_{T:t-1} \sim q_\theta(X_{T:t-1})} \left [ \Tilde{H}(X_{t-1},X_t)  \right ] \\ 
       & + \beta \, \mathbb{E}_{X_{T:0} \sim q_\theta(X_{T:0})} \left [ H(X_0) \right ]  + C,
\end{align*}
where we have exchanged the sums over $X_{T:t}$ with the expectation over $X_{T:t} \sim q_\theta(X_{T:t}) $. Finally, we can arrive at Eq.~\ref{eq:loss} by multiplying the whole objective with the temperature $\Tau$.

\subsection{Gradient of the Joint Variational Upper Bound}
\label{app:gradient}
When $X_{0:T}$ is discrete, the gradient of the Joint Variational Upper Bound has to be decomposed into two parts via the application of the chain rule. The gradient through exact expectations (see Sec.~\ref{sec:methods}) can be computed by backpropagating through $q_\theta(X_{t-1}|X_t)$ directly. To backpropagate through the expectations over $X_{T:t-1} \sim q_\theta(X_{T:t-1})$ we use the log derivative trick with a baseline for variance reduction.

As an example, we show the gradient through the entropy term which is given by:

\begin{align*}
    \nabla_\theta \mathbb{E}_{X_{T:t} \sim q_\theta(X_{T:t})} \left [ S(q_\theta(X_{t-1}|X_t)) \right ] & = \mathbb{E}_{X_{T:t} \sim q_\theta(X_{T:t})} \biggl [ \left [ S(q_\theta(X_{t-1}|X_t))  - b_S \right ] \nabla_\theta \log{q_\theta(X_{T:t})} \biggr ] \\
    & + \mathbb{E}_{X_{T:t} \sim q_\theta(X_{T:t})} \left [ \nabla_\theta S(q_\theta(X_{t-1}|X_t)) \right ], 
\end{align*}
where the baseline $b_S = \mathbb{E}_{X_{T:t} \sim q_\theta(X_{T:t})} \left [ S(q_\theta(X_{t-1}|X_t)) \right ]$ is used for variance reduction.
Here, the first term is the REINFORCE gradient through the expectation and the second term is the gradient which flows through the exact expectation of the entropy.

\subsection{Exact expectations of mean-field distributions}
\label{app:exact_expectations}

\paragraph{Exact Expectation of Energy Functions.}
In the following, we will derive the exact expectations from Sec.~\ref{sec:methods}.
Suppose we have a mean-field distribution $p(X) = \prod_{i = 1}^N p(X_i)$
and want to calculate $\mathbb{\mathop{E}}_{X \sim p(X)} \left [  Q_{ij} X_i X_j \right ]$, where $X_i \in \{0,1\}$, then when $i \neq j$

\begin{align*}
    \mathbb{\mathop{E}}_{X \sim p(X)} \left [ Q_{ij} X_i X_j \right ] &= \sum_{X} p(X) \, Q_{ij}\, X_i \,X_j  = \sum_{X}  \prod_{i = 1}^N p(X_i) \, Q_{ij} \,X_i \,X_j \\
    &= \sum_{X_i} \sum_{X_j} p(X_i)\, p(X_j)\, Q_{ij} \, X_i \,X_j = Q_{ij} \, p(X_i = 1) \,p(X_j = 1) := Q_{ij} \, p_i \, p_j.
\end{align*}
In the third equality, we have integrated out the probabilities that do not depend on $X_i$ and $X_j$ and in the fourth equality we have used that only terms where $X_i = X_j = 1$ will remain. For $i = j$ it can be similarly shown that $\mathbb{\mathop{E}}_{X \sim p(X)} \left [ Q_{ii} X_i X_i \right ] = Q_{ii} \, p_i$, by using that $X_i^2 = X_i$.

\paragraph{Exact Expectation of Bernoulli Noise Distribution.}
\label{app:noise_energy}
As explained in Sec.~\ref{sec:methods} for the binary case $X_i \in \{ 0,1\}$ the Categorical Noise Distribution can be reduced to a Bernoulli Noise Distribution.

The logarithm of this noise distribution can be written as:

\begin{equation}
\begin{aligned}
    \Tilde{H}(X_t, X_{t-1}) & := \sum_i (1-X_{t,i}) \biggl [ (1-X_{t-1,i}) \log{ (1- \beta_t)} + X_{t-1,i} \log{\beta_t} \biggr ]\\
    & + X_{t,i} \biggl [ (1- X_{t-1,i}) \log{ (1-\beta_t )} + X_{t-1,i} \log{\beta_t} \biggr]
\end{aligned}
\label{eq:app:noise_energy}
\end{equation}

Similarly, as done before we can calculate the expectation with respect to a mean-field distribution $p(X_t)$ exactly, where we arrive at
\begin{align*}
    p(X_t|{X_{t+1}}) \, p(X_{t-1}|X_t) \, \Tilde{H}(X_t, X_{t-1}) &= p(X_t|{X_{t+1}}) \, \sum_i (1-X_{t,i}) \biggl [ p_{t-1,i} \log{ (1- \beta_t)}  + (1-p_{t-1,i}) \log{\beta_t} \biggr]\\
    & + X_{t,i} \biggl [ (1- p_{t-1,i}) \log{ (1-\beta_t )} + p_{t-1,i} \log{\beta_t} \biggr],
\end{align*}  
where we define $ p_{t-1,i} := p(X_{t-1,i} = 1|X_t) $. The expectation over $p(X_t|{X_{t+1}}) $ cannot be computed in closed form because $p_{t-1,i}$ depends nontrivially on $X_t$.

\paragraph{Exact Expectation of the Entropy.}
For the exact expectation of the entropy we can similarly show that:

\begin{align*}
    S(p(X)) & = - \sum_X p(X) \log p(X) = \sum_{X} \prod_i p(X_i) \sum_j \log p(X_j) \\
    & = - \sum_{X} \sum_j \prod_i p(X_i) \log p(X_j) \\ 
    &= - \sum_j \sum_{X_j} p(X_j) \log p(X_j) = \sum_j \biggl [ p_j \log p_j - (1-p_j) \log (1-p_j) \biggr ]
\end{align*}

\subsection{Equivalence of one step DiffUCO and EGN}
\label{app:EGNisOneStepDiffUCO}

In diffusion models, the last step of the forward diffusion process typically corresponds to sampling independent random noise. For a one-step diffusion model as considered in Sec.~\ref{sec:egn_sec} this implies that $p(Z|X)$ does not depend on $X$.
With this insight it follows that: $$ C = \sum_{X, Z } q_\theta(X|Z) \, q(Z)  \, \log{p(Z|X)} = \sum_{Z} \, q(Z)  \, \log{p(Z)}.$$
This shows that $C$ is independent of $\theta$.
\section{Additional Experiments}
\label{app:add_exp}
\subsection{Experiments on the Gset Dataset}
We furthermore evaluate our method on the Gset Maxcut dataset, where models are trained and validated on Erdős–Rényi graphs and then evaluated out-of-distribution on Gset graphs \citep{Gset}. We train our method on Erdős–Rényi (ER) graphs between 200 to 500 nodes that are generated with a uniformly sampled edge probability $p \in \{ 0.05, 0.3\}$. The neural baseline methods ECORD \citep{ECORD}, ECO-DQN \citep{ECO-DQN}, RUNCSO \citep{toenshoff_graph_2020}, and ANYCSP \citep{anycsp} use graphs with the same edge probability but train on graphs with vertex sizes of 500, 200, 100, and 100 respectively. All methods are validated on ER graphs with 500 vertices. Results are also compared to a greedy construction algorithm (Greedy) and semidefinite programming (SDP) \citep{SDP} which is a well-known approximation algorithm on MaxCut.
Results are shown in Tab.~\ref{tab:Gset}, where the difference to the best-known cut value is shown. We compare to results reported as in ANYSCP \citep{anycsp}. Here, Gset graphs are grouped according to their vertex size $|V|$ and then on each graph models run with 20 parallel processes for a time limit of 180 seconds. Results show that our model achieves the second-best results and ANYCSP achieves the best results in all groups.
However, a comparison between DiffUCO and ANYCSP cannot be considered to be fair since DiffUCO is a solution generation method whereas ANYCSP is a search method. The former category of methods is trained to generate solutions without iteratively improving a solution, which differentiates them from search methods. Therefore, search-based methods such as ANYCSP have an advantage in the setting of sampling 20 solutions in parallel within the time limit of 180 seconds per graph. 
Furthermore, the comparison between neural methods in this setting may 
not be fair because the test dataset is out-of-distribution (OOD) and the training datasets of the neural methods also differ in some cases. For a more detailed discussion on OOD generalization experiments, we refer to App.~\ref{app:OOD}.

\subsection{Out of distribution generalization}
\label{app:OOD}
We present in Tab.~\ref{tab:OOD_results} OOD results on Barabási-Albert (BA) graphs for both the MaxCut and the Minimum Dominating Set (MDS) problems. Our approach is trained on BA-small graphs and then evaluated on datasets comprising larger graphs, including BA-large, BA-huge, and BA-giant with 800-1200, 1200-1800, and 2000-3000 nodes, respectively. In this assessment, we benchmark our method against Gurobi with a generous time constraint of 300 seconds per graph. We observe remarkably good generalization capabilities on these datasets and even outperform Gurobi on BA-large, BA-huge, and BA-giant MaxCut at the given time limit. However, OOD results have to be treated with great caution as it is well known \citep{jegelka1, jegelka2} that OOD generalization on graph problems does not only depend on the method but also the GNN architecture and the node-to-degree distribution of the graphs. Our method was by no means designed for robust OOD generalization, which would be an attractive avenue for future work.

\subsection{Effect of different token sizes in CE-ST}
\label{app:ST_ablation}
In Fig.~\ref{fig:Rebuttal} (Left) we show how the solution quality and the inference speed of our model changes on the RB-100 MIS dataset when we vary the token size $k$ within CE-ST. Here, the relative error $\epsilon_\mathrm{rel}$ is plotted on the left axis and the evaluation time in seconds on the right axis. We see that $k$ does not influence the solution quality of the model and the inference time has an optimum at $k = 8$. Because some operations within our implementation of CE-ST scale exponentially with $k$ the evaluation time increases at $k > 8$. By further optimizing the CE-ST code in this respect, the negative time effects of these operations might be reduced.

\subsection{DiffUCO learns to generate samples that are more likely under the forward diffusion process}
In Fig.~\ref{fig:Rebuttal} (Right) we show that our model indeed learns a diffusion process.
To show that samples are related to the diffusion process we plot the evolution of the unnormalized forward process likelihood $\mathbb{E}_{X_{0:T} \sim q_\theta(X_{0:T})} [ \log p(X_{0:T})]$ averaged over model samples throughout training. 
As this measure increases during training, this indicates that the model indeed learns to generate samples that are related to the diffusion process, i.e. it shows that the generative process is increasingly reconstructing the forward diffusion process.

\subsection{Factoring in training time}
In Tab.~\ref{tab:n_query} we quantify the number of $n_{\mathrm{Query}}$, where $t_{\mathrm{train}} +  n_{\mathrm{Query}} \cdot t_{\mathrm{eval}} < t_\mathrm{Gurobi}$. Under the assumption that the solution quality of Gurobi and the model are the same, this quantity is supposed to show at which amount of samples it is more useful to use our model instead of Gurobi.
Our results show that the number of $n_{Query}$ on some problems is quite large, while on others it is comparably low.
On MaxCut, for example, Gurobi takes a lot of time to obtain a good solution quality, and due to the low connectivity of the BA graphs our method does not take a lot of time for training. In contrast to that $n_{Query}$ is very large on MaxCl because, as we have mentioned in App B.6, the MaxCl experiments are the most expensive due to the high connectivity of the complementary RB graphs.

\begin{table*}[h!]
\centering
\small
\setlength\tabcolsep{2pt}
\begin{adjustbox}{width=.75\textwidth}
{\renewcommand{\arraystretch}{1.8}
\begin{tabular}{c | c c  | c c| c c |c}
 CO problem type & MaxCut & & MIS & & MDS & & MaxCl \\
\hline
 Dataset & BA-small & BA-large & RB-small & RB-large & BA-small & BA-large & RB-small \\
 \hline
$n_{\mathrm{Query}}$ & 482 & 197 & 8.596 & 48.216 & 348.830 & 31.525 & 1.752.059

\end{tabular}

}
\end{adjustbox}
\caption{The number of $n_{\mathrm{Query}}$, where $t_{\mathrm{train}} +  n_{\mathrm{Query}} \cdot t_{\mathrm{eval}} < t_\mathrm{Gurobi}$. Here, $t_{\mathrm{train}}$ is the training time of DiffUCO, $t_{\mathrm{eval}}$ is DiffUCO`s evaluation time per graph and $t_\mathrm{Gurobi}$ is the evaluation time per graph of Gurobi.}
\label{tab:n_query}
\end{table*}

\begin{table*}[h!]
\centering
\small
\setlength\tabcolsep{2pt}
\begin{adjustbox}{width=.65\textwidth}
\begin{tabular}{c c c c c }
Method & $|V| = 800$ & $|V| = 1K$ & $|V| = 2K$ & $|V| \geq 3K$  \\
\hline
Greedy (r) & 411.44 & 359.11 & 737.00 & 774.25 \\
SDP (r) & 245.44 & 229.22 & N/A  & N/A \\
RUNCSP (r) & 185.89 & 156.56 & 357.33  & 401.00 \\
ECO-DQN (r) & 65.11 & 54.67 & 157.00 & 428.25 \\
ECORD (r)& 8.67 & 8.78 & 39.22 & 187.75 \\
ANYCSP (r)& $\mathbf{1.22}$ & $\mathbf{2.44}$  & $\mathbf{13.11}$ & $\mathbf{51.63}$\\
\hline
DiffUCO: CE-ST$_8$ & 4.11 & 6.33 & 31.67 & 116.75 \\

\end{tabular}
\end{adjustbox}
\caption{Comparison to MaxCut results form \citet{anycsp}. Models are evaluated according to the average deviation from the best-known cut size on the Gset dataset. The graphs are grouped according to their node size $|V|$. (r) indicates that the results are taken from \citet{anycsp}.}
\label{tab:Gset}
\end{table*}

\begin{table*}[h!]
\centering
\small
\setlength\tabcolsep{2pt}
\begin{adjustbox}{width=.7\textwidth}
\begin{tabular}{c c c c c }
 &  & MDS Size $\downarrow$ &  & \\
 \hline
Method & BA-small & BA-large & BA-huge & BA-giant  \\
\hline
Gurobi - $t_{lim} = 300 s$ & $\mathbf{27.91}$ & $\mathbf{105.05}$ &  $\mathbf{153.89}$ & $\mathbf{253.53}$ \\
DiffUCO: CE-ST$_8$ & 28.31 & 106.08 & 157.46 & 259.02 \\
\hline
&  & MaxCut Size $\uparrow $& &   \\
\hline
Gurobi - $t_{lim} = 300 s$ & $\mathbf{735.67}$ & 2945.43 & 4414.29  & 7319.47  \\
DiffUCO: CE-ST$_8$ & 731.72 & $\mathbf{2949.02}$ & $\mathbf{4444.47}$ & $\mathbf{7390.99}$  \\

\end{tabular}
\end{adjustbox}
\caption{Out-of-distribution (OOD) results on MDS and MaxCut. DiffUCO is trained on BA-small and then the average set size is evaluated on three larger OOD datasets BA-large, BA-huge with an average of 2000 nodes, and BA-giant with an average of 3000 nodes. Gurobi is evaluated at a time limit of 300s per graph. The best method is marked as bold.}
\label{tab:OOD_results}
\end{table*}

\begin{figure*}[h]
    \centering
    \begin{minipage}{0.48\textwidth}
    \centering
    \setlength\tabcolsep{2pt}
  \includegraphics[width=1.\linewidth]{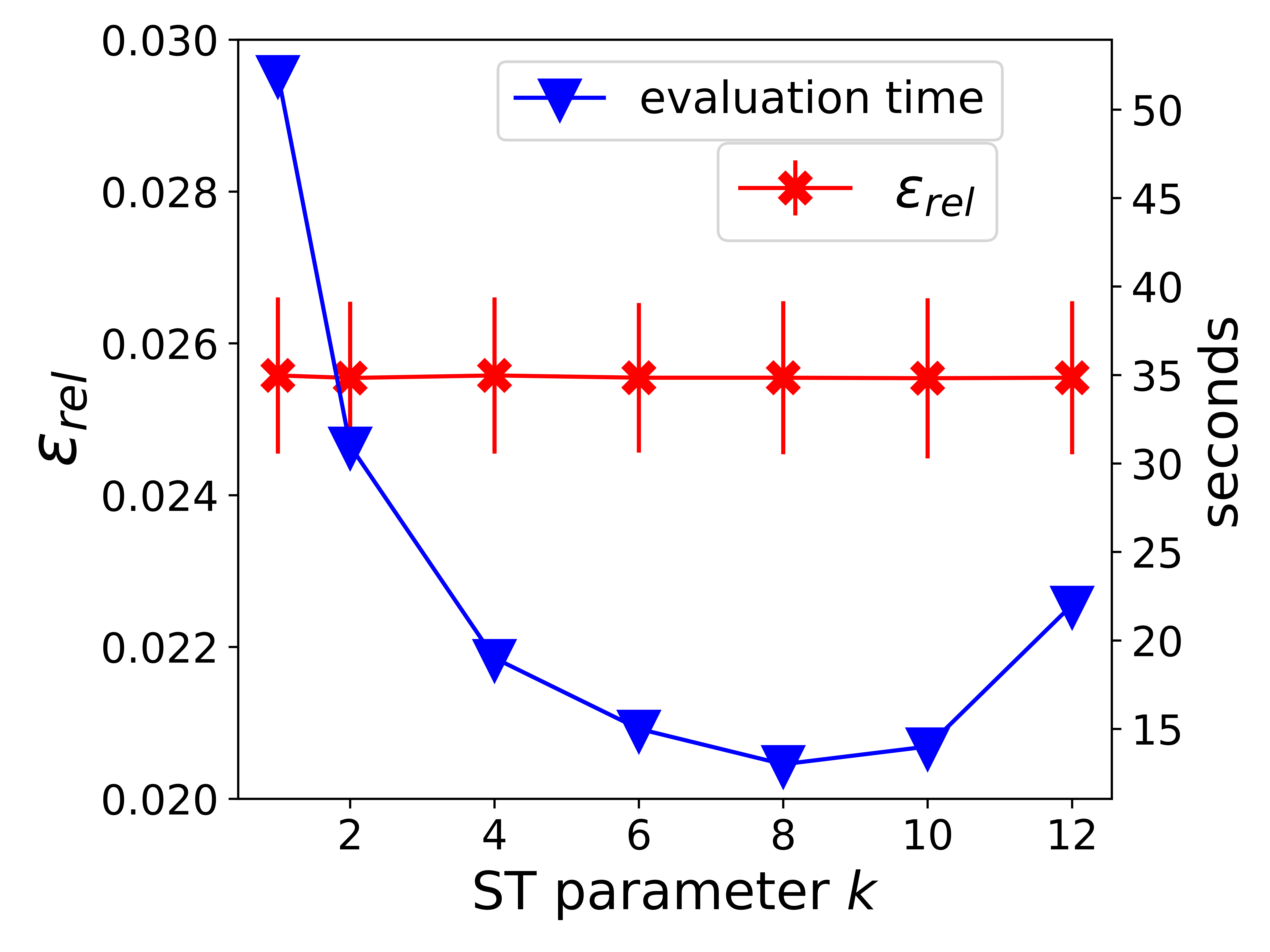}
  \end{minipage}
      \begin{minipage}{0.48\textwidth}
    \centering
    \setlength\tabcolsep{2pt}
  \includegraphics[width=1.\linewidth]{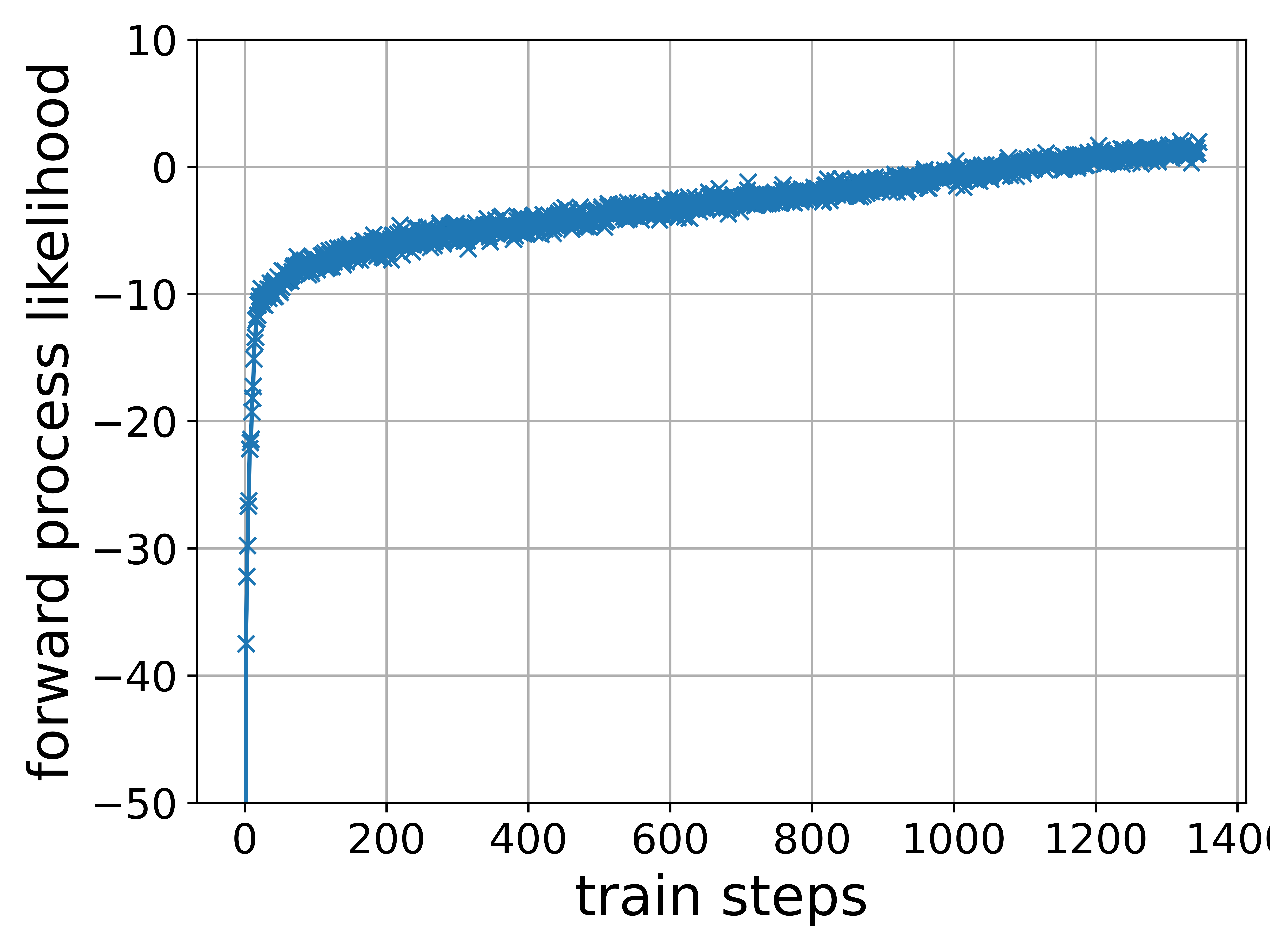}
  \end{minipage}
\vspace{-2ex}
\caption{Left: Evaluation of the solution quality of CE-ST over $k$ on the RB-small MIS dataset. The plot shows that the relative error does not change when $k$ is increased, whereas the evaluation time decreases up to $k = 8$. As some operations scale exponentially with $k$ we observe that the evaluation time increases at $k > 8$. Right: The unnormalized forward process likelihood  $\mathbb{E}_{X_{0:T} \sim q_\theta(X_{0:T})} [ \log p(X_{0:T})]$ over the number of training steps on the RB-100 MIS dataset. The model is trained at a constant temperature. The plot shows that the model learns to generate samples that become more likely under the forward diffusion process. }
\label{fig:Rebuttal}
\end{figure*}

\section{Experimental Details}

\subsection{Metrics}
\label{app:metrics}
\paragraph{Relative Error}
The relative error is defined as $$ \widehat{\epsilon}_{rel} = \frac{ | E_{\mathrm{opt}} - \widehat{E}_{\mathrm{model}}|}{|E_{\mathrm{opt}}|},$$ where $E_{\mathrm{opt}}$ is the optimal energy and  $\hat{E}_{\mathrm{model}}$ is the average energy of the model.

Similarly, $\epsilon^*_{rel}$ is the best relative error, where the best out of eight samples $E^*_{\mathrm{model}}$ instead of the average model energy $\hat{E}_{\mathrm{model}}$ is used.

\paragraph{Approximation Rate}
The best approximation rate is defined as $$ \mathrm{AR}^* = \frac{E^*_{\mathrm{model}}}{E_{\mathrm{opt}}},$$ where $E_{\mathrm{opt}}$ is the optimal energy and  $E^*_{\mathrm{model}}$ is the energy of the best out of eight samples from the model.

For approximation rate and relative error, lower values are better.
Our experiments are always averaged over three independent seeds.

\subsection{Evaluation diffusion steps}
\label{app:evaluation_diff}
In some experiments (see Sec.~\ref{sec:benchmarks} and see Sec.~\ref{sec:ablations}) we evaluate DiffUCO on the test dataset by applying more diffusion steps than we have used during training.

We do this in the following way:\\
When the diffusion model is trained with $T$ diffusion steps, at diffusion step $t$ the diffusion model is additionally conditioned on a one-hot encoding of this time step. Therefore, if we instead want to run DiffUCO with $n \times T$ diffusion steps, we repeat each time step $n$ times. So during evaluation, at each time step $t \in \{ 1, ... , n \times T \}$ a $X_t$ is sampled from the diffusion model and reprocessed at the next time step. At each time step the model is time conditioned on the value $\left \lfloor \frac{t+1}{n} 
 \right \rfloor$. Here, $\lfloor \cdot \rfloor$ is an integer operation.

\subsection{Datasets}
\label{app:datasets}
\paragraph{RB-100 MIS dataset}
The RB-100 MIS dataset is generated by specifying generation parameters $n, k^\prime$, and $p$. 
With $n$ the number of cliques, i.e. a set of fully connected nodes, and with $k^\prime$ the number of nodes within the clique are specified. $p$ serves as a parameter that regulates the interconnectedness between cliques. The lower the value of $p$ the more connections are randomly drawn between cliques. If $p = 1$ there are no connections between the cliques at all.
To generate the RB-100 dataset with graphs of an average node size of 100, we generate graphs with $n \in \{ 9,15 \}$, $k^{'} \in \{ 8,11 \}$, and $p \in \{ 0.25, 1 \}$. 
\paragraph{RB-200 MVC dataset}
On the RB-200 dataset $k \in \{ 9 , 10 \}$ and $n \in \{ 20,25\}$ , and $p \in \{ 0.25, 1 \}$. We use 2000 graphs for training, 500 for validation, and for each p we use 100 graphs for testing.

\paragraph{RB-small and RB-large}
On the RB-small dataset $k \in \{ 5, 12 \}$ and $n \in \{ 20, 25\}$ and graphs that are smaller than 200 nodes or larger than 300 nodes are resampled. On BA-large $k \in \{ 20,25 \}$ and $n \in \{ 40, 55\}$ and graphs that are smaller than 800 nodes or larger than 1200 nodes are resampled. For both of these datasets $p \in \{ 0.3, 1 \}$.

\subsection{Architecture}
\label{app:architecture}

We use a simple GNN Architecture, where in the first step input each node feature is transformed by a linear layer with $n_h$ neurons. These node embeddings are then multiplied with a weight matrix with $n_h$ neurons followed by a sum aggregation over the neighborhood. Additionally, a skip connection on each node is added. Afterward, a NodeMLP processes the aggregated nodes together with the skip connection. After $n$ message passing steps each node embedding is fed into a final three-layer MLP which computes the probabilities of each solution variable $X_i$. Layernorm \citep{ba_layernorm_2016} is applied after every MLP layer, except for the last layer within the final MLP. We always use $n_h = 64$ in all of our experiments.

\subsection{Hyperparameters}
\label{app:hyperparams}
A table with all hyperparameters on each dataset is given in Tab.~\ref{tab:Hyperparameters}. $M_\omega$ is the batch size of different CO problem instances and $M_{KL}$ is the batch size with which the Joint Variational Upper Bound is estimated.
RAdam \citep{RADAM} is used as an optimizer. We clip the gradients to a norm of $1.0$. 

In experiments on the RB dataset, except for MaxCl, we additionally use $5$ random node features to increase the expressivity of the GNN in critical cases. This is necessary to improve the performance on the RB dataset at low $p$ values.

\begin{table*}[h!]
\centering
\small
\setlength\tabcolsep{2pt}
\begin{adjustbox}{width=.8\textwidth}
{\renewcommand{\arraystretch}{1.8}
\begin{tabular}{c c c c c c c c c}
  Dataset  & lr & $T_{\mathrm{start}}$ & GNN layers & diffusion steps & $N_\mathrm{anneal}$ & noise distr. & $M_\omega$ & $M_{KL}$\\
\hline
\hline
  RB-large MIS & 0.002 & 0.3 & $7$ & 4 & 3000 & annealed noise distr. & 18 & 5
\\
\hline
 RB-small MIS & 0.002 & 0.4 & $8$ & 6 & 4000 & annealed noise distr. & 30 & 10
\\
\hline
 RB-small MaxCl & 0.002 & 0.5 & $8$ & 5 & 3000 & cat. noise distr. & 20 & 8
 \\
 \hline
 BA-large MaxCut & 0.002 & 0.2 & $4$ & 6 & 1000 & annealed noise distr. & 20 & 10
 \\
  \hline
 BA-small MaxCut & 0.002 & 0.2 & $8$ & 4 & 2000 & annealed noise distr. & 20 & 10
 \\
  \hline
 BA-large MDS & 0.003 & 0.3 & $8$ & 3 & 2000 & annealed noise distr. & 20 & 10
 \\
  \hline
 BA-small MDS & 0.003 & 0.3 & $8$ & 5 & 2000 & annealed noise distr. & 20 & 10
 \\
  \hline
 RB-200 MVC & 0.001 & 0.4 & $8$ & 4 & 4500 & annealed noise distr. & 30 & 10
 \\
       

\end{tabular}

}
\end{adjustbox}
\caption{Table with all hyperparameters on each dataset.}
\label{tab:Hyperparameters}
\end{table*}

\subsection{Computational Requirements}
\label{app:resources}
Experiments on RB graphs are quite memory expensive, because of the high connectivity of these graphs. Especially experiments on MaxCl are expensive because the energy aggregation on the complementary graph leads to even higher connectivity and therefore to a large computational graph. On RB-large MIS experiments are for example conducted on one A100 GPU with 80 GB. Experiments on BA graphs have much fewer computational requirements because the connectivity of these graphs is quite low.
Comparing the training time of our method to baselines taken from \citet{gflow_2023} is difficult as the authors do not report the training time nor the computational resources of their method or the other methods used in their paper. They also do not provide a list of hyperparameters such as the number of epochs for any method, which makes the reproducibility of their results difficult. We hypothesize that LTFT might need shorter training and fewer computational resources on problems like MIS and MaxCl because, for these CO problems, the authors design problem-specific Markov Decision Processes that are very efficient for these kinds of problems.
However, this makes their approach in contrast to our approach less general because these tricks for example cannot be applied to MDS and MaxCut. Therefore, we expect LTFT to have long training times on MaxCut and MDS. This assumption is supported by the long evaluation times of LTFT on these CO problems as reported in Tab.~\ref{tab:MIS} and Tab.~\ref{tab:MaxCl}. On MaxCut their evaluation time is more than 21 times longer and on MDS the evaluation time is more than 32 times longer compared to our method. If we compare the training time of our method to VAG-CO on the RB-200 dataset, we can state that our method takes 2d 2h for training and VAG-CO takes 3d 7h for training. Other baselines like EGN and EGN-Anneal converge quite fast but do not achieve a competitive solution quality due to a lack of expressivity \citep{VAG-CO}.
In EGN-Anneal the training duration highly depends as in all annealing-based methods on the annealing schedule. The more annealing steps are used the longer the training time. However, here the solution quality with longer annealing also saturates due to a lack of expressivity.

\subsection{Time Measurement}
\label{app:time}
All time measurements were conducted on an A100 NVIDIA GPU.
We only include the time of the forward pass and the time needed to perform CE in the time measurement. Therefore, we do not include the time that is necessary to load the graphs from the dataloader. 
For the forward pass, we measure the time on jitted functions. Different states from the diffusion model can in principle be computed in a parallelized manner. Therefore, we measure the time sequentially on each state and then compute the average of this measured time.
Self conducted Gurobi evaluations on MaxCut are run on a Intel Xeon Platinum 8168 @ 2.70GHz CPU with 24 cores.

\subsection{Code}
The code for this research project is based on jax \citep{jax2018github}.

\end{document}